\documentclass[preprint,12pt]{elsarticle}
\usepackage{lineno,hyperref}
\usepackage{multirow}
\usepackage{makecell}
\usepackage{amsmath,amsfonts}
\usepackage{algorithm}
\usepackage[noend]{algpseudocode}
\usepackage{array}
\usepackage[caption=false,font=normalsize,labelfont=sf,textfont=sf]{subfig}
\usepackage{textcomp}
\usepackage{stfloats}
\usepackage{url}
\usepackage{verbatim}
\usepackage{graphicx}
\usepackage{booktabs}
\usepackage{xcolor}

\usepackage[table]{xcolor}
\definecolor{myyellow}{RGB}{255, 255, 180}

\definecolor{blue}{rgb}{0,0,0}

\definecolor{red}{rgb}{0,0,0}

\usepackage{makecell} % 
\usepackage{multirow} % 
\usepackage{threeparttable}
\usepackage{comment}

\usepackage{pdflscape}
\usepackage{booktabs}
\usepackage{tikz}
\newcommand{\excuparrow}{\tikz[baseline=-0.5ex]\fill[green!70!black] (0,0) -- (0.8ex,0) -- (0.4ex,0.8ex) -- cycle;}
\newcommand{\excdownarrow}{\tikz[baseline=-0.5ex]\fill[red!70!black] (0,0.8ex) -- (0.8ex,0.8ex) -- (0.4ex,0) -- cycle;}

\def\BibTeX{{\rm B\kern-.05em{\sc i\kern-.025em b}\kern-.08em\TeX}}

\begin{document}
\begin{frontmatter}
\title{Heterogeneous Model Fusion for Privacy-Aware Multi-Camera Surveillance via Synthetic Domain Adaptation}
%\title{ Integrating Heterogeneous Object Detectors within a Federated Learning Architecture  }
\author[1]{Peggy Joy Lu\corref{cor1}}
\ead{peggylu@cs.ccu.edu.tw}

\author[2,3]{Wei-Yu Chen}
%\ead[chen]{wychen@narlabs.org.tw}

\author[1]{Yao-Tsung Huang}

\author[2]{Vincent Shin-Mu Tseng}
%\ead[tseng]{vtseng@nycu.edu.tw}

\affiliation[1]{organization={Department of Computer Science and Information Engineering, National Chung Cheng University},
    city={Chiayi},
    country={Taiwan}}

\affiliation[2]{organization={Department of Computer Science, National Yang Ming Chiao Tung University},
    city={Hsinchu},
    country={Taiwan}}

\affiliation[3]{organization={National Center for High-Performance Computing},
    city={Taichung},
    country={Taiwan}}

\cortext[cor1]{Corresponding author}

\begin{comment}
    
\author{
    \IEEEauthorblockN{Peggy Joy Lu\IEEEauthorrefmark{1}, Wei-Yu Chen\IEEEauthorrefmark{2}\IEEEauthorrefmark{3}, Vincent Shin-Mu Tseng\IEEEauthorrefmark{2}}
    \IEEEauthorblockA{
        \IEEEauthorrefmark{1}National Chung Cheng University, Chiayi, Taiwan \\
        Email: peggylu@cs.ccu.edu.tw
    }
    \IEEEauthorblockA{
        \IEEEauthorrefmark{2}National Yang Ming Chiao Tung University, Hsinchu, Taiwan \\
        Email: \{waue0920.cs07,vtseng\}@nycu.edu.tw
    }
    \IEEEauthorblockA{
        \IEEEauthorrefmark{3}National Center for High-Performance Computing, Taichung, Taiwan \\
        Email: wychen@narlabs.org.tw
    }
}
\end{comment}            

%\maketitle

\begin{abstract}

We propose HeroCrystal (\underline{He}te\underline{ro}geneous Model Fusion for Privacy-Aware Multi-\underline{C}ame\underline{r}a Surveillance via \underline{Sy}n\underline{t}hetic Domain \underline{A}daptation), a novel privacy-preserving framework for multi-camera domain-adaptive object detection, addressing challenges such as data privacy, class imbalance, and heterogeneous architectures. 

Our framework consists of three key stages. In the \textbf{Generated Stage}, we introduce a one-shot, target-aware diffusion-based generation module that learns visual style from a single target-domain image while leveraging prompt-based control to synthesize specific object instances. Unlike conventional style transfer-based methods that require large target datasets and ignore semantic-level discrepancies, our approach enables privacy-preserving augmentation to reduce ethical concerns, and introduces controllable rare object generation to mitigate long-tailed category degradation. In the \textbf{Federated Stage}, we employ probabilistic Faster R-CNN on the client side to improve localization accuracy, and a dynamic model contrastive strategy to suppress domain-specific bias. The server side performs model fusion across heterogeneous architectures without accessing raw data. Finally, in the \textbf{Distilled Stage}, we propose an inconsistent categories integration algorithm to resolve label inconsistency and architecture heterogeneity across clients.

Extensive experiments on multiple cross-domain detection benchmarks demonstrate that our method outperforms existing multi-source domain adaptation and federated learning baselines under multi-class, privacy-preserving settings. Our method improves mAP by +2.1\% over prior privacy-preserving approaches and achieves a new state-of-the-art mAP of 33.4\%, highlighting the effectiveness of HeroCrystal in enabling practical multi-camera AI surveillance systems.

The source code is publicly available at \url{https://github.com/ccuvislab/HeroCrystal}.

\end{abstract}

%\begin{IEEEkeywords}
\end{frontmatter}

\section{Introduction}
\label{sec:intro}

Smart city surveillance systems increasingly rely on networks of distributed cameras deployed across diverse environments such as roads, campuses, and intersections. Building a robust object detector in such scenarios requires training on diverse visual conditions, which motivates the use of domain adaptation (DA) techniques. However, most existing multi-source domain adaptive object detection (MSDAOD) methods \cite{yao2021multi, wu2022target, PMT2024} assume centralized access to both source and target data during model training. This assumption is often impractical in real-world deployments, where aggregating raw data from all camera sources raises significant privacy concerns and faces regulatory restrictions. 

To solve the privacy issue, Federated learning (FL) \cite{FedAvg} offers a promising solution, allowing model training across multiple clients without sharing raw data. In the context of surveillance, the federated scenario has been recently formulated as MSDAOD under privacy constraints \cite{lu2022fusion}, where labeled source datasets reside on separate clients and the unlabeled target-domain data remains isolated at the server. While this setup better aligns with privacy-preserving requirements, it introduces new challenges in handling domain discrepancies, label inconsistencies, and model fusion from heterogeneous clients.

FL architectures enforce strict data isolation, each client can only access its local data, and no raw data can be shared between domains. Importantly, this constraint protects not only the privacy of the source data but also that of the target domain. As a result, the learning task on each client effectively becomes a source-only domain adaptation problem. Therefore, we propose a source-only \textbf{probabilistic Faster R-CNN}, which is better suited for object detection tasks where accurate localization is essential. Moreover, training detectors on locally heterogeneous data often leads to overly domain-specific features that generalize poorly. To reduce this overfitting, we further introduce a \textbf{dynamic contrastive learning strategy} that encourages feature alignment between local client features and the global model. 

In real-world federated deployments, \color{blue} distinct environments and object distributions across cameras often result in heterogeneous model architectures caused by category-inconsistent datasets, posing significant challenges for server-side model fusion. \color{black}Traditional MSDAOD or privacy-preserving MSDAOD methods typically restrict the detection task to a single category, e.g. 'car', which simplifies the problem but limits practical applicability. In our work, we propose our \textbf{inconsistent categories integration} and \textbf{model fusion algorithm} to address the issue of label inconsistency across clients, thereby enabling multi-category object detection under realistic and scalable federated settings.

\begin{figure}[ht]
\centering

   \includegraphics[width=.7\linewidth]{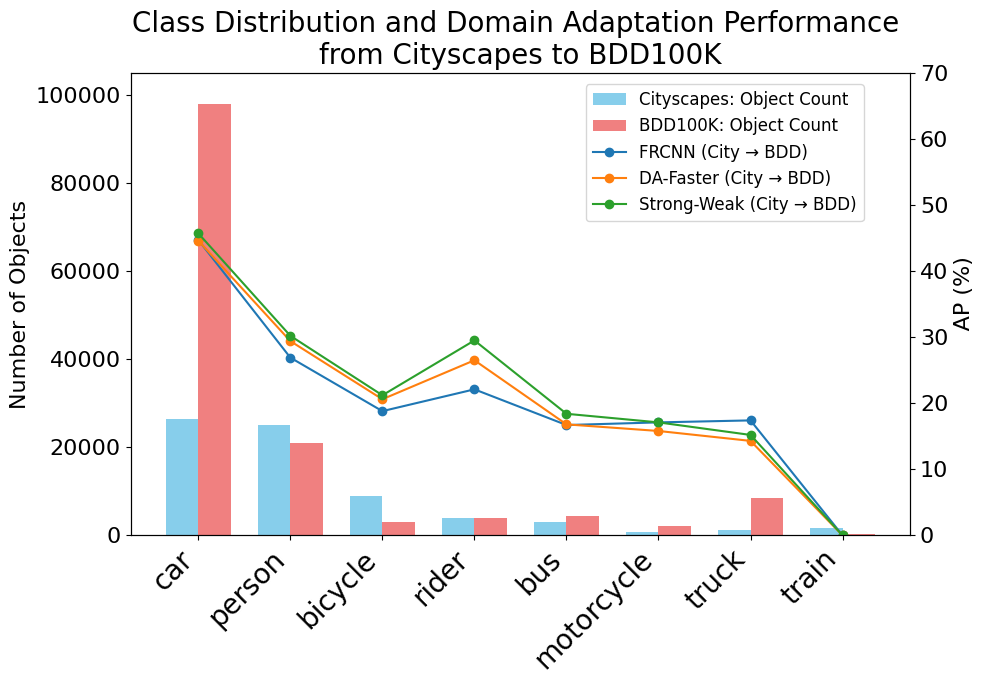}

  \caption{ Comparison of the object distribution between the Cityscapes \cite{cordts2016cityscapes} and BDD100K \cite{Yu2020BDD100K} datasets, and the corresponding cross-domain detection performance (mAP) from Cityscapes to BDD100K using FRCNN \cite{faster-RCNN}, DA-Faster \cite{chen2018domain}, and Strong-Weak \cite{saito2019strong}}
  \label{fig:class_distribution}
\end{figure}

\color{blue}Our scenario can be regarded as an MSDAOD setting under privacy-preserving constraints, which means that each client is trained as a source-only model. Previous works have attempted to extract domain-invariant representations from multiple client models \cite{lu2023ICIP} or employ pseudo-label learning strategies \cite{FedCoin}, but their improvements remain limited when strict privacy restrictions prevent direct access to target data. Other methods commonly adopted in domain adaptation translate source images into the visual appearance of the target domain using GAN-based \cite{bousmalis2017pixelda} or diffusion-based \cite{huang2024blenda} style transfer. \color{black} However, both GAN and diffusion-based methods typically require access to a large number of target images to capture domain characteristics, which raises the privacy concerns for target data. To protect the privacy of both source and target data, we propose a \textbf{one-shot, target-aware diffusion generation module} that synthesizes target-style images. Our method learns visual style and scene layout from a single target-domain image, and uses an off-the-shelf object detector to automatically generate pseudo-labels of the synthetic data. 

Besides privacy concern, we observe that long-tailed categories face significant performance degradation in domain adaptation. As shown in Figure ~\ref{fig:class_distribution}, rare categories such as \textit{train}, \textit{truck}, and \textit{motorcycle} consistently suffer from lower detection accuracy, mainly due to their scarcity in the source data. \color{blue}This issue becomes more severe in FL scenarios as multiple clients capture distinct visual domains with highly imbalanced and partially overlapping category distributions. \color{black} Although existing style transfer approaches can effectively align global appearance statistics such as color and texture, they lack the ability to control semantic properties of the scene, such as object placement, density, and co-occurrence. Therefore, we design our generation module to synthesize specific object instances within target-style scenes through prompt-based control, which requires no additional annotations or datasets and enables balanced augmentation of long-tail categories. As a result, our method not only preserves data privacy but also provides class-aware augmentation that boosts recognition accuracy for long-tailed categories.

\begin{table}[hb]
    \centering
    \makebox[\textwidth][c]{%

\resizebox{1.2\textwidth}{!}{%
    \begin{threeparttable}
        \caption{Comparison between our works and existing methods across various functionalities. \color{black}}
        \label{table:features}
            \begin{tabular}{ccccccc}
                \hline
            \textbf{Methods} &

            \textbf{\makecell{Unbalanced \\ Data}}  & 
            \textbf{\makecell{Multi- \\ Source}} & 
            \textbf{\makecell{Privacy- \\ Preserving}}  & 
            \textbf{\makecell{Multi-Class \\  Prediction}} & \textbf{\makecell{Heterogeneous \\ Architectures}} & 
            \textbf{\makecell{ Diffusion-based \\ Generation}} \\  
            
            \hline
Single-source DA & $\bigcirc$ &  &  & $\bigcirc$ &  & $\triangle$\\
Multi-source DA & $\bigcirc$ & $\bigcirc$ &  &  &  & \\
Federated Learning & $\bigcirc$ & $\bigcirc$ & $\bigcirc$ &  &  & \\
\textbf{ours} & $\bigcirc$ & $\bigcirc$ & $\bigcirc$ & $\bigcirc$ & $\bigcirc$ & $\bigcirc$\\

            \hline
            \end{tabular}
            \vspace{0.5em}
\begin{tablenotes}
\item[*] \textnormal{$\bigcirc$}: \textnormal{fully supported; } $\triangle$: \textnormal{partially supported.}

\item[*] \textbf{Unbalanced Data} indicates that the training datasets contains unbalanced category distribution.
\item[*] \textbf{Multi-Class Prediction} refers to the model's ability to detect a large number of distinct classes across domains.
\end{tablenotes}

    \end{threeparttable}
    } }% resize
\end{table}

%table1
To highlight the functional scope of our contributions, Table~\ref{table:features} compares key functionalities across similar works. Existing single-source DA methods typically overlook privacy concerns and multi-source diversity, although some employ style transfer for target appearance adaptation, they often rely on centralized data access. For multi-source DA methods, while leveraging diverse source domains, commonly require access to both source and target data, which compromises privacy. In addition, due to inconsistent label sets across sources, they tend to simplify the task by detecting only the most common category. 
% reviewer fix - <R1-M5> - waue 0922
\color{blue}
Traditional FL frameworks preserve data privacy but assume homogeneous architectures and label spaces, whereas in real-world surveillance, category-inconsistent datasets lead to heterogeneous detection heads and consequently heterogeneous model architectures. Our method effectively addresses these three aspects of heterogeneity by supporting heterogeneous model fusion, multi-source learning over inconsistent label sets, and style-aware synthetic data augmentation for imbalanced data within a privacy-preserving federated framework.
\color{black}

\begin{comment} 
Traditional federated learning frameworks preserve data privacy but assume homogeneous model architectures, making them unsuitable for heterogeneous, multi-class object detection. In contrast, our method uniquely supports unbalanced training data, multi-source learning, heterogeneous model fusion, and diffusion-based generation, all within a privacy-preserving federated framework.
\end{comment}

% reviewer fix - <R1-M5> - waue 0922

To summarize, our main contributions are:
\begin{enumerate}
\item To preserve the privacy across distributed cameras, we propose a federated architecture and adopt one-shot target-aware generation that requires only a single image from the target domain for personalization.

% reviewer fix - <R1-M5> - waue 0922
\color{blue}
\item To address the heterogeneity in data annotation, data distribution, and model labeling, we propose an innovative inconsistent categories integration mechanism combined with a model fusion algorithm.
\color{black}
% \item To address the difficulty of fusing heterogeneous models with inconsistent number of categories, we introduce an inconsistent categories integration mechanism combined with model fusion algorithm.
% reviewer fix - <R1-M5> - waue 0922

\item To enrich long-tail categories and capture domain-specific object semantics beyond global style, we develop a prompt-driven diffusion-based generation module that explicitly synthesizes target-style images with controllable object classes.
\end{enumerate}

\section{Related Works}
\subsection{Heterogeneity in Federated Learning}

Federated learning (FL) has emerged as a privacy-preserving distributed machine learning paradigm, enabling clients to collaboratively train a global model without sharing their local private data~\cite{FedAvg}. A key challenge in real-world FL is data heterogeneity, as data across clients is typically highly non-IID due to differences in data sources and user behaviors~\cite{zhao2018federated}. Such non-IID characteristics can severely affect the convergence speed and final performance of the global model.

To mitigate the negative effects of non-IID data, researchers have proposed various strategies. Some works focus on improving server-side aggregation and fusion algorithms, such as FedProx~\cite{li2020fedprox}, which introduces a proximal term to limit divergence from the global model, and more advanced causal-based fusion methods like FuseFL~\cite{tang2024fusefl}. Other directions include client clustering and personalized FL frameworks~\cite{tan2022towards}. When client data exhibits significant domain shifts, federated transfer learning provides a promising solution by enabling knowledge transfer across domains, which is especially important for tasks such as visual object detection, exemplified by methods like FedVision~\cite{liu2020fedvision}.

The core challenge addressed in this work arises from a more extreme form of data heterogeneity than typical non-IID settings. While existing FL studies on non-IID data often assume clients share similar underlying feature spaces, our setting directly confronts clients from entirely different domains with distinct data distributions, label spaces, and annotation types.

% reviewer fix - <R1-M2> - waue 0922
\subsection{Multi-Source Domain Adaptation for Object Detection}

Domain adaptation (DA) techniques aim to address the performance degradation caused by distribution shifts between training (source) and testing (target) data~\cite{wang2018deep}. In many scenarios, multiple labeled source domains are available, giving rise to multi-source domain adaptation (MSDA), which integrates knowledge from multiple sources for superior generalization on the target domain~\cite{mansour2009domain}. Early MSDA methods focused on learning a shared feature space by minimizing moment discrepancies or through adversarial learning to encourage domain-invariant representations~\cite{MDAN}. 
\color{blue} However, recent studies have pointed out that domain adaptive object detection (DAOD) tasks often suffer from noisy annotations, which can significantly hinder adaptation robustness \textcolor{red}{~\cite{liu2022robust}}. This highlights the importance of designing MSDA approaches that not only align across domains but are also resilient to annotation noise in practical detection scenarios.

To circumvent direct source data access, a notable line of work explores \emph{source-free domain adaptation} (SFDA)~\cite{VS2023, liu2023periodically, Yoon2024}, wherein a source-pretrained model adapts to the target domain without any source data. SFDA methods employ strategies such as self-training with pseudo-labels, teacher-student learning, and uncertainty-aware loss formulations to mitigate domain shift solely from unlabeled target data. However, SFDA often assumes a single source domain, necessitates large volumes of unlabeled target samples, and relies on the consistent architectural setting across domains, limiting scalability and applicability.

Our work distinguishes itself by operating within a FL framework, aggregating knowledge from multiple heterogeneous clients without exposing source data. Unlike SFDA, our approach accommodates heterogeneous annotation spaces and model architectures through a novel \emph{inconsistent categories integration} mechanism. Furthermore, by employing a one-shot, target-style image generation strategy, we drastically reduce dependency on abundant target samples, essential for privacy-critical environments and long-tailed category adaptation.

\color{black}
\subsection{Style-Aware Generative Domain Adaptation}
A fundamental challenge in supervised learning is the performance drop caused by domain shift. Among various domain adaptation (DA) techniques, style transfer has emerged as an effective paradigm for enabling knowledge transfer across domains with significant appearance differences in illumination or weather~\cite{hoffman2018cycada}. Early works like PixelDA~\cite{bousmalis2017pixelda} leveraged GANs~\cite{goodfellow2020gan} to transform labeled source images into target-like counterparts at the pixel level, while others like ADDA~\cite{tzeng2017adda} moved adversarial learning into the feature space to align source and target encoders.

To further improve fidelity, subsequent GAN-based works enhanced generator backbones to mitigate content distortion. %~\cite{hicsonmez2020ganilla}. 
More recently, research has shifted toward diffusion models, which surpass GANs in generation stability and detail preservation. To mitigate semantic distortion, StyleDiffusion~\cite{wang2023stylediffusion} introduces a two-stage pipeline guided by CLIP~\cite{radford2021clip} embeddings to separate content from style, while BlenDA~\cite{huang2024blenda} introduces a diffusion-based blending mechanism that combines generated samples with source images to form intermediate-domain images for smoother adaptation. To reduce computational overhead, recent strategies focus on lightweight fine-tuning of pre-trained models using methods like LoRA~\cite{hu2022lora} or on training-free frameworks that inject style during inference by modulating attention mechanisms~\cite{huang2025attenst}.

Despite these advancements, most style transfer methods remain constrained to deterministic image-to-image mappings that primarily alter global visual style while preserving the source image's content structure. We propose a prompt-driven diffusion framework that explicitly learns to model and generate not only the visual style but also the underlying semantic layout of the target domain. This allows our method to synthesize diverse and realistic training images with controllable object classes and spatial arrangements, offering a more effective solution for domain adaptation in complex scenarios.

\color{black}

\section{Proposed Method}

\begin{figure*}[htb]
\centering
% \makebox[\textwidth][c]{ % [R2-C1] removed oversized makebox; fit within textwidth
\includegraphics[width=\textwidth]{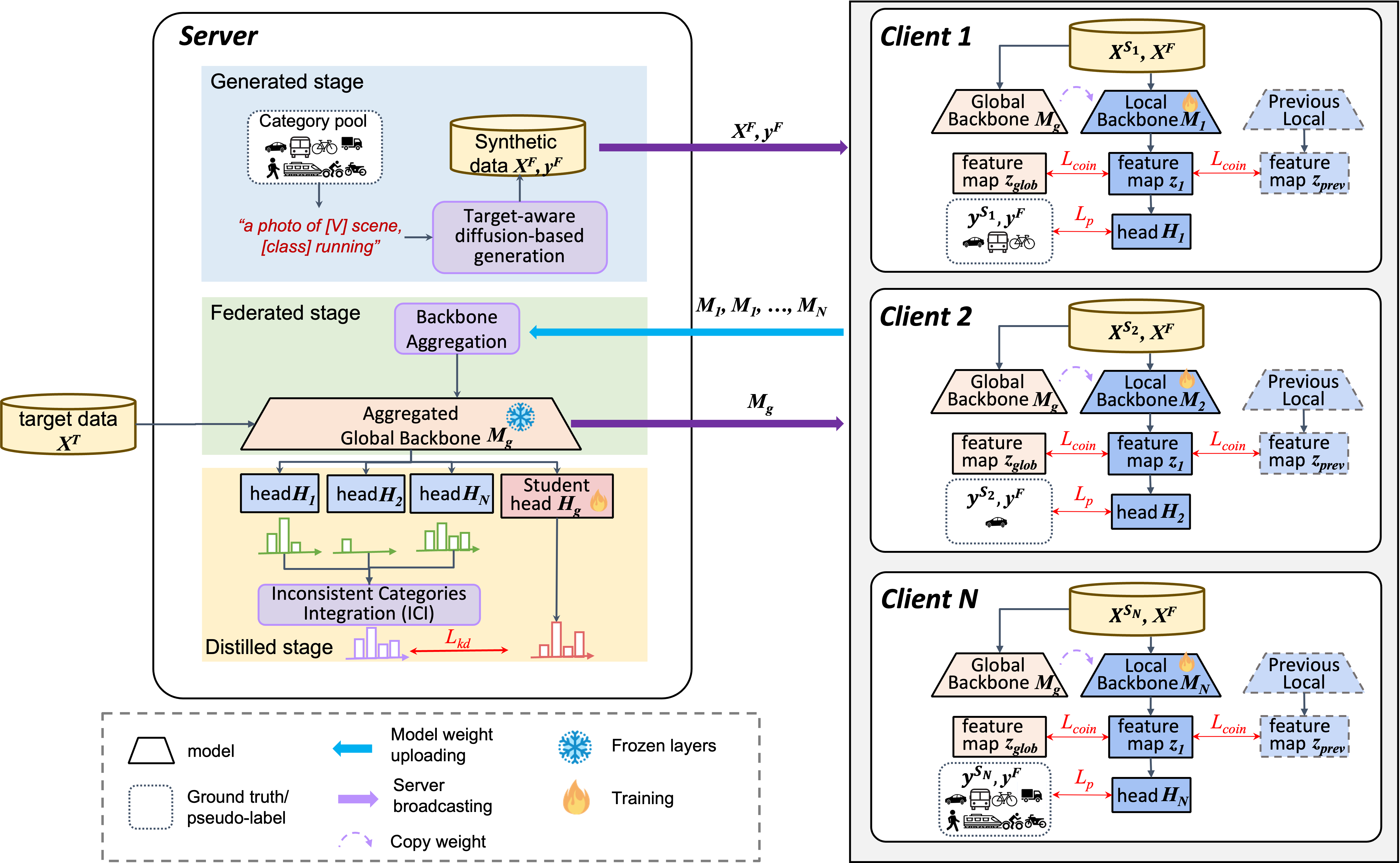} % [R2-C1] 1.2\linewidth -> \textwidth
	% } % [R2-C1] close of removed makebox
\caption{\textcolor{red}{Overview of the system architecture.}}
\label{fig:architecture}
\end{figure*}

\subsection{Architecture}

Our system architecture, as illustrated in Figure~\ref{fig:architecture}, follows a FL setup with $N$ clients and a central server. Each client owns a localized labeled dataset that remains private throughout the training process. The overall learning pipeline consists of three stages: the \textbf{generated stage}, the \textbf{federated stage}, and the \textbf{distilled stage}.

In the \textbf{generated stage} (Sec. \ref{sub:augmented}), the server utilizes a target-aware diffusion-based generation module to synthesize target-style data with pseudo-label. In particular, we can specify the desired foreground categories to generate by sampling class names from a predefined category pool and incorporating them into the text prompts. These synthetic samples, denoted as $(X^F, y^F)$, are later distributed to all clients as auxiliary data.

During the \textbf{federated stage}, each client receives a global backbone $M_g$ and performs local training using both its own labeled source data and the synthetic target-style data. Based on these data, the client updates its local backbone $M_i$ and detection head $H_i$ by minimizing two losses: a probabilistic detection loss $L_p$ (Sec. \ref{subsub:PFRCNN}) and a contrastive alignment loss $L_{con}$ (Sec.~\ref{subsub:dynamic_moon}). After local updates, clients upload their updated backbones to the server, which aggregates them into a new global backbone (Sec. \ref{sub:bakcbone_agg}). This federated training process is repeated over multiple rounds to gradually refine the global backbone.

Finally, in the \textbf{distilled stage}, the server freezes the aggregated global backbone $M_g$ and distills the output knowledge from all client heads $\{H_i\}_{i=1}^N$ into a student head $H_g$. To address the issue of heterogeneous architectures and varying category sets across clients, we adopt the inconsistent categories integration (ICI) mechanism (Sec.~\ref{sub:KD}) to combine predictions before applying the distillation loss $L_{kd}$. The resulting head $H_g$, together with the aggregated global backbone $M_g$, served as the final detector.

\begin{table}[htbp]
\centering
\caption{Notation Definitions}
\label{tab:notation}

\resizebox{\textwidth}{!}{%

\begin{tabular}{ll}

\toprule
\textbf{Notation} & \textbf{Definition} \\
\midrule
\multicolumn{2}{l}{\textbf{Category 1: Image, Label and embedding}} \\

$X^T, X^S, X^F$ & Image from different domains. (T: target, S: source, F: Synthetic) \\
$y^F$ & Pseudo-label corresponding to synthetic image \\$c_j^i$ & Class prediction of proposal $j$ from client $i$ \\
$p_j^i$ & Pseudo-label $j$ predicted by client $i$ \\
$P_l$ & Integrated pseudo-label set for class $l$ \\
$J_i$ & Number of proposals from source $i$ \\
$\{(X_k^i, B_k^i, C_k^i)\}_{k=1}^{n_i}$ & A mini-batch of image, bounding boxes, and class label from client $i$ \\
$L$ & Set of categories in the target domain \\
$(b_x, b_y, b_w, b_h)_j^i$ & Coordinates of proposal $j$ from client $i$ \\
$z$, $z_{glob}$, $z_{prev}$ & Feature embeddings (local, global, previous) \\
$\mathbf{u}_t$ & Latent representation at timestep $t$ \\
$\boldsymbol{\phi}_v$, $\boldsymbol{\phi}_g$ & Prompt embedding for a target-specific instance/ generic scene\\
$\epsilon_\theta$ & Predicted noise from denoising UNet \\
\midrule
\multicolumn{2}{l}{\textbf{Category 2: Model Components}} \\
$M_i^r \ (i{=}1,\dots,N, r{=}1,\dots,R)$ & Local model of client $i$ at round $r$\\
$M_i\ (i{=}1,\dots,N)$, $M_g$ & Models of local clients and the global, respectively \\
$H_i\ (i{=}1,\dots,N)$, $H_g$ & Detection head of $i$ client and the global fused head, respectively \\

\midrule
\multicolumn{2}{l}{\textbf{Category 3: Parameters}} \\
$\lambda$ & Dynamic weighting coefficient for contrastive loss \\
$\beta$ & Trade-off parameter between reconstruction and prior loss\\

$\tau$ & Temperature parameter in contrastive loss \\
$\alpha_t$ & Noise scheduling controlling parameter at timestamp $t$ \\

\midrule
\multicolumn{2}{l}{\textbf{Category 4: Training Configuration}} \\
$E$ & Number of local training epochs \\
$R$ & Total number of communication rounds \\

\midrule
\multicolumn{2}{l}{\textbf{Category 5: Loss Functions}} \\
$\mathcal{L}_p$ & Probabilistic detection loss \\

$\mathcal{L}_{kd}$ & Distillation loss for head fusion \\
$\mathcal{L}_{moon}$ & Contrastive loss using global model as positive sample\\
$\mathcal{L}'_{moon}$ & Contrastive loss using global model as negative sample \\
$\mathcal{L}_{con}$ & Final contrastive loss with dynamically weighted fusion \\
$\mathcal{L}_{recon}$ & Reconstruction loss for style matching \\
$\mathcal{L}_{prior}$ & Prior preservation loss for general prompt \\
$\mathcal{L}_{gen}$ & Generation loss \\

\bottomrule
\end{tabular}
}
\end{table}

\subsection{Symbols and Definitions}
To ensure clarity and consistency throughout the paper, we summarize the key notations used in our framework in Table~\ref{tab:notation}. The notations are organized into five main categories: (1) image, label, and embedding representations, which describe multi-domain inputs, pseudo-labels, and proposal-related features; (2) model components, including local and global backbones and detection heads involved in the federated architecture; (3) training parameters, such as weights for contrastive losses, temperature values, and noise scheduling coefficients used in the generative module; (4) configuration settings that specify the number of local training epochs and communication rounds; and (5) loss functions, covering objectives related to detection, contrastive alignment, generation, and prompt preservation. These notations provide a unified reference for all algorithmic designs and mathematical formulations presented in the paper.

\subsection{Stage 1: Target-aware Diffusion-based Generation}
\label{sub:augmented}

\begin{figure}[htb]
\centering
\subfloat[]{  \includegraphics[width=.65\linewidth]{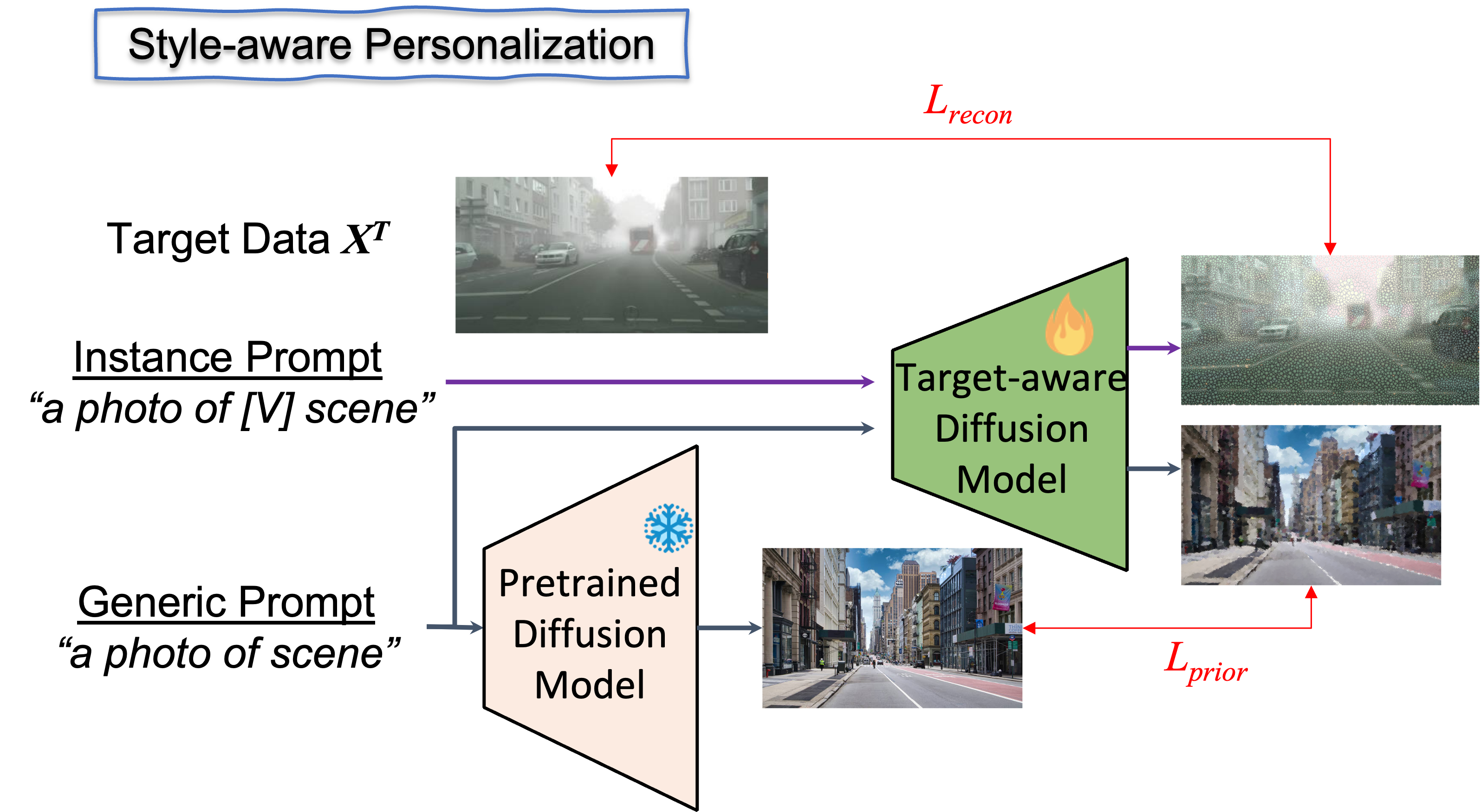}} \\
\subfloat[]{  \includegraphics[width=\linewidth]{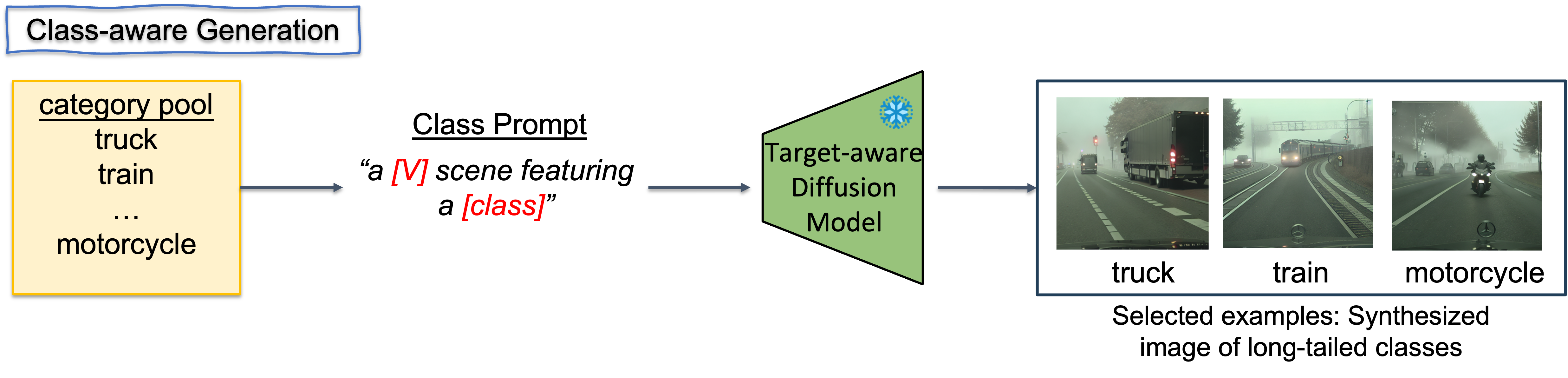}}

  \caption{  (a) Style-aware personalization: adapting the diffusion model to the target domain using instance- and generic prompts. (b) Class-aware generation: synthesizing target-style images guided by prompts that combine domain style and specific object categories; long-tailed categories are shown here for illustration.}
  \label{fig:dreambooth}
\end{figure}

%\subsection{Long-tail Category-aware Target-domain Stylization via Diffusion Models}

As illustrated in Figure~\ref{fig:class_distribution}, common categories (e.g., cars, persons) dominate the dataset, while long-tailed categories (e.g., buses, trains) are significantly underrepresented, resulting in degraded detection performance under domain adaptation. To address this imbalance, we introduce a \textit{target-aware diffusion-based generation module} that selectively synthesizes target-style images with controllable object semantics, thereby enhancing domain generalization for rare categories. To achieve this, we adopt the conditional diffusion model \cite{ruiz2023dreambooth} and tailor it to a one-shot setting that requires only a single image from the target domain, ensuring both personalization and data privacy. The overall pipeline is illustrated in Figure~\ref{fig:dreambooth}, which consists of (a) \textbf{Style-aware personalization} for training a domain-style diffusion model and (b) \textbf{Class-aware generation} for synthesizing specific object by control the prompt. 

During the \textbf{Style-aware personalization}, the training process is guided by two types of prompts and their corresponding supervision to adapt the model toward the target domain. Specifically, an instance prompt (e.g., “a photo of [V] scene”), where [V] is a special learned token capturing the visual style of the target domain, is used to guide the model to generate images consistent with the reference style. Accordingly, we apply a reconstruction loss:

\begin{equation}
\mathcal{L}_{\text{recon}} = \mathbb{E}_{t, \mathbf{u}_0, \epsilon} \left[ \left\| \epsilon - \epsilon_\theta(\mathbf{u}_t, t, \boldsymbol{\phi}_v) \right\|^2 \right]
\label{eq:recon}
\end{equation}

where \( \mathbf{u}_t = \sqrt{\alpha_t} \, \mathbf{u}_0 + \sqrt{1 - \alpha_t} \, \epsilon \) is the noisy latent at timestep \( t \),  
\( \epsilon \sim \mathcal{N}(0, \mathbf{I}) \),  
\( \alpha_t \in [0, 1] \) is a noise scheduling parameter that controls the trade-off between the clean latent \( \mathbf{u}_0 \) and the injected noise \( \epsilon \),  
\( \boldsymbol{\phi}_v \) is the embedding of the instance prompt containing the learnable [V] token,  
and \( \epsilon_\theta \) is the noise prediction UNet conditioned on both \( t \) and \( \boldsymbol{\phi}_v \).

In addition, a generic prompt (e.g., “a photo of scene”) is adopted to maintain the model’s general visual prior and prevent overfitting. A prior preservation loss is introduced accordingly:

\begin{equation}
\mathcal{L}_{\text{prior}} = \mathbb{E}_{t, \mathbf{u}_0, \epsilon} \left[ \left\| \epsilon - \epsilon_\theta(\mathbf{u}_t, t, \boldsymbol{\phi}_g) \right\|^2 \right]
\label{eq:prior}
\end{equation}

where \( \boldsymbol{\phi}_g \) is the embedding of a generic prompt representing a scene description without object semantics.

The final training objective is a weighted sum of both losses:
\begin{equation}
\mathcal{L}_{\text{gen}} = \mathcal{L}_{\text{recon}} + \beta \cdot \mathcal{L}_{\text{prior}}
\label{eq:total}
\end{equation}
where \( \beta \) balances the target adaptation and prior retention.

\begin{figure}[htbp]
\centering
\subfloat[]{\includegraphics[width=0.24\linewidth,height=2.5cm]{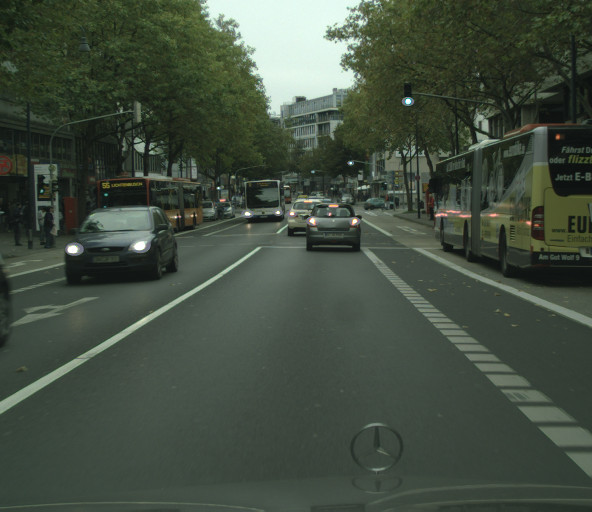}}\hfill
\subfloat[]{\includegraphics[width=0.24\linewidth,height=2.5cm]{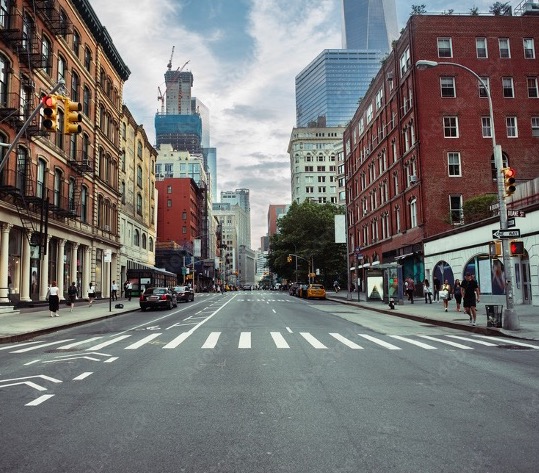}}\hfill
\subfloat[]{\includegraphics[width=0.24\linewidth,height=2.5cm]{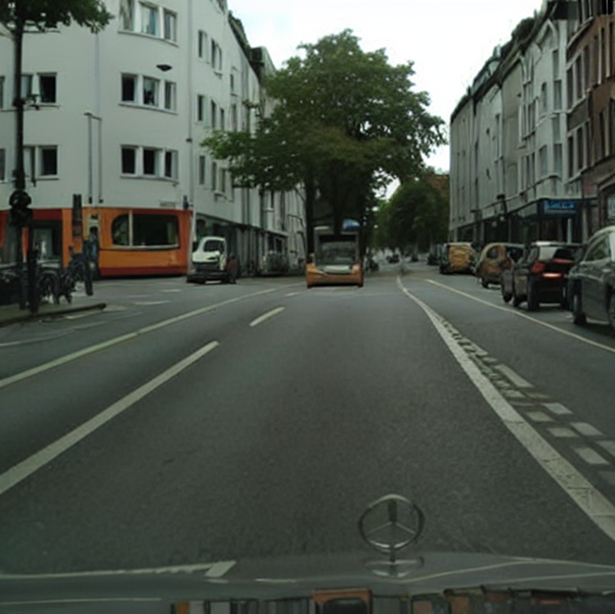}}\hfill
\subfloat[]{\includegraphics[width=0.24\linewidth,height=2.5cm]{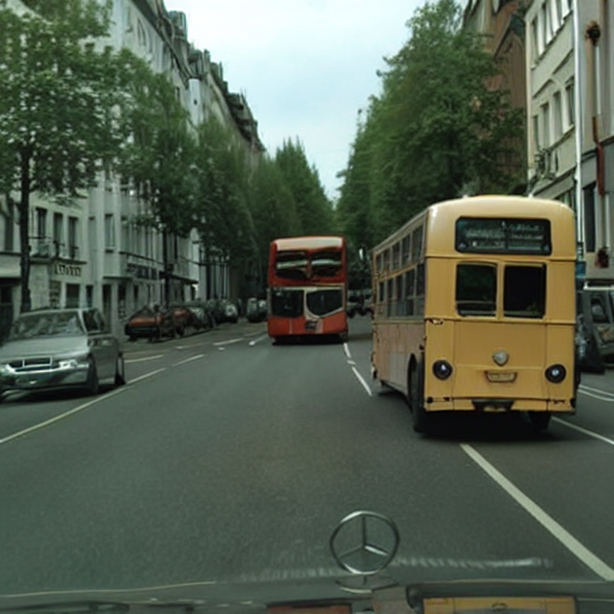}}
\caption{\textcolor{red}{In the generated stage, we employ target-aware data generation to synthesize specific objects under a specified style.
(a) Target-style reference image from the target domain;
(b) Image generated from the same prompt without style adaptation;
(c) Image generated with the learned target style;
(d) Image generated with both the target style and a specified object (e.g., bus).}}
\label{fig:generated_example}
\end{figure}

In the \textbf{Class-aware generation} phase, target-style images are synthesized by composing prompts that integrate the learned [V] token with specific object categories (e.g., “a photo of [V] scene with a bus”). This enables the model to synthesize diverse scenes that reflect the visual style of the target domain while embedding specific object semantics, as visualized in Figure~\ref{fig:generated_example}.
A single reference image guides the generation to preserve contextual backgrounds of common objects while explicitly introducing rare object instances, enhancing exposure to long-tail semantics. To facilitate downstream training, an off-the-shelf detector assigns pseudo-labels to the synthesized instances.

As shown in Table~\ref{table:dataset_stats}, the original datasets exhibit long-tailed distributions, where certain categories such as truck and train are severely underrepresented. To mitigate this imbalance, we generated 100 target-style images with a specific object for each target dataset, while also allowing selective augmentation for long-tailed categories as analyzed in the ablation study (Sec.~\ref{sub:generate_single_ablation}).\footnote{Note that the rider class is excluded from generation due to its strong visual overlap with bicycle and motorcycle, in order to maintain category balance and avoid over-representation.}

Overall, this generation mechanism supports domain adaptation in two key aspects: (1) it ensures privacy by requiring only one target image; and (2) it selectively enriches rare categories by synthesizing targeted examples under the target domain’s style, helping to mitigate semantic shifts in long-tailed object detection.

\begin{table*}[htbp]
\centering
\caption{
Summary of datasets used in the two experimental scenes. 
Scene (a): Sim10k, KITTI, Foggy $\rightarrow$ Cityscapes. 
Scene (b): Cityscapes, KITTI $\rightarrow$ BDD100K. 
Only diffusion-generated target-style data [O] are used for training. 
Original target-domain datasets are used only for evaluation (validation split) or for one-shot style reference. 
\textbf{[O]} denotes \textit{one-shot diffusion}, i.e., 100 synthetic images per class generated from a single reference image. 
\textit{Validation datasets (e.g., Cityscapes-val and BDD100K-val) refer specifically to the validation splits of the original datasets, used exclusively for evaluation.}
}
\label{table:dataset_stats}
\resizebox{\textwidth}{!}{%
\begin{tabular}{llllrrrrrrrr}
\toprule
\textbf{Scene} & \textbf{Dataset} & \textbf{Role} & \textbf{Used For} & \textbf{Car} & \textbf{Person} & \textbf{Bicycle} & \textbf{Rider} & \textbf{Bus} & \textbf{\makecell{Motor\\cycle}} & \textbf{Truck} & \textbf{Train} \\
\midrule
\multirow{5}{*}{(a)} 
& Sim10k & Source & Training & 58710 & -- & -- & -- & -- & -- & -- & -- \\
& KITTI & Source & Training & 25932 & 21610 & 0 & 4322 & 0 & 0 & 1094 & 511 \\
& Foggy Cityscape & Source & Training & 26500 & 25000 & 9000 & 4000 & 3000 & 800 & 1200 & 1600 \\
& Cityscapes [O] & Synthetic Target & Training & 100 & 100 & 100 & 100 & 100 & 100 & 100 & 100 \\
& Cityscapes-val & Target Val Set & Evaluation & 5648 & 1241 & 1100 & 657 & 345 & 365 & 415 & 458 \\
\midrule[0.2pt]
\multirow{4}{*}{(b)} 
& KITTI & Source & Training & 25932 & 21610 & 0 & 4322 & 0 & 0 & 1094 & 511 \\
& Cityscapes & Source & Training & 26500 & 25000 & 9000 & 4000 & 3000 & 800 & 1200 & 1600 \\
& BDD100K [O] & Synthetic Target & Training & 100 & 100 & 100 & 100 & 100 & 100 & 100 & 100 \\
& BDD100K-val & Target Val Set & Evaluation & 5052 & 8426 & 674 & 886 & 654 & 1025 & 1852 & 78 \\
\bottomrule
\end{tabular}%
}
\end{table*}

\subsection{Stage 2: Federated Training with Contrastive Alignment}
\label{sub:federated}

In our FL architecture, the local training process incorporates two approaches: Probabilistic Faster R-CNN (Sec. \ref{subsub:PFRCNN}) and Dynamic Model Contrastive Learning (Sec. \ref{subsub:dynamic_moon}). The former introduces uncertainty modeling into the localization head to better handle ambiguous object boundaries, while the latter aims to suppress domain-specific features by encouraging consistency across local models.After each round, the server aggregates client backbones using a model fusion algorithm to obtain a domain-invariant global representation (Sec.~\ref{sub:bakcbone_agg}).

\subsubsection{Probabilistic Faster R-CNN }
\label{subsub:PFRCNN}

Two-stage object detectors separate region proposal and object recognition into distinct stages, offering greater flexibility and modularity. This design allows the proposal mechanism to adapt independently of the final detection head, which is widely used in domain adaptive object detection. Faster R-CNN~\cite{faster-RCNN} is one of the most commonly used two-stage architectures in domain adaptive object detection. It jointly performs classification (\textit{cls}) and bounding box regression (\textit{bbox}) tasks.

In its original formulation, Faster R-CNN applies cross-entropy loss for classification and $L1$ loss for regression, introducing uncertainty only in the classification branch. To better handle false negatives (FNs) without access to target domain annotations, we adopt the regression head using in \cite{lu2023ICIP}, which introduces uncertainty modeling into bounding box predictions.\footnote{A detailed analysis of false negative reduction is provided in Sec.~\ref{sub:FN_reduce_ablation}.} Each bounding box \( B = (b_x, b_y, b_w, b_h) \) consists of four coordinates, where each \( b_i \) is modeled as a one-dimensional Gaussian distribution \( \mathcal{N}(\mu_i, \sigma_i^2) \), capturing both the predicted value and its uncertainty.

The regression loss is then computed as the negative log-likelihood of the ground truth box ${B}^{GT}$ under the predicted distribution:

\begin{equation}
\label{eq:pfrcnn}
\begin{aligned}
\mathcal{L}_{bbox}= \frac{1}{J_{bbox}} \sum_{i} \mathbb{I}_{fg}(t_i)\mathcal{H}({B}^{GT}_{i}, {B}_{i}) \\
\cong \frac{1}{J_{bbox}} \sum_{i} \mathbb{I}_{fg}(t_i) \log (\mathcal{N}({B}^{GT}_{i}; {\mu}_i, {\sigma}^2_i))
\end{aligned}
\end{equation}

where ${B}_i$ is the predicted bounding box, ${\mu}_i$ and ${\sigma}^2_i$ denote the mean and variance of its coordinates, $\mathbb{I}_{fg}(t_i)$ is an indicator for foreground proposals, and $J_{bbox}$ is the total number of proposals. This probabilistic formulation enhances the model’s robustness by expressing uncertainty in localization. The overall loss becomes:
\vspace{-0.3em}
\begin{equation}
\label{eq:faster}
\mathcal{L}_{p} = \mathcal{L}_{cls} + \mathcal{L}_{bbox}
\end{equation}

\subsubsection{Dynamic Model Contrastive Learning }
\label{subsub:dynamic_moon}

Moon \cite{moon} performs model-level contrastive learning by comparing representations produced by different models. It assumes that the global model captures more generalizable features than local models, and thus treats the global model as the positive sample and the local model as the negative sample. Based on this assumption, the contrastive loss in Moon is defined as:
 
\begin{equation}
\label{equ:moon}
%\begin{multiline}
\mathcal{L}_{moon} = -\log \left( \frac{\exp\left(\text{sim}(z, z_{\text{glob}})/ \tau\right)}{\exp\left(\text{sim}(z, z_{\text{glob}})/\tau\right) + \exp\left(\text{sim}(z, z_{\text{prev}})/\tau\right)} \right)
%\end{multiline}
\end{equation}
\vspace{0.5em}
where $\mathbf{z}$, $\mathbf{z}_{glob}$ and $\mathbf{z}_{prev}$ are the feature representation from different backbones. The function $\text{sim}(\mathbf{u}, \mathbf{v}) = \frac{\mathbf{u}^\top \mathbf{v}}{|\mathbf{u}| |\mathbf{v}|}$ represents the cosine similarity between vectors $\mathbf{u}$ and $\mathbf{v}$, and $\tau$ is a temperature parameter.

\begin{figure}[htb]

\centering

   \includegraphics[width=.90\linewidth]{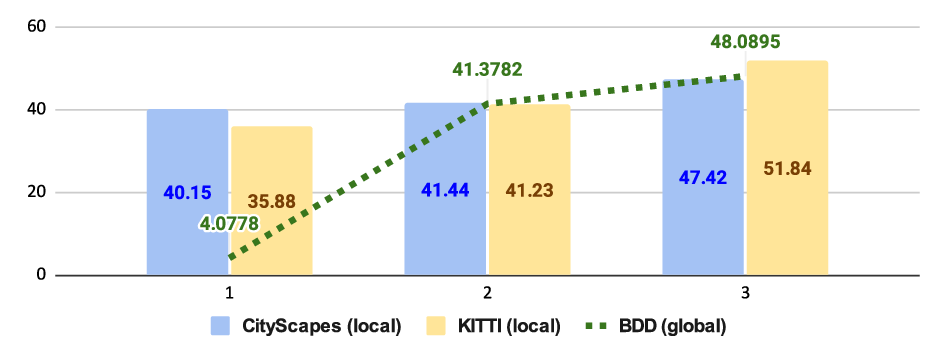}

  \caption{ The trend of APs for the \textit{CK$\rightarrow$B} setting \color{blue} by using only the standard FedAvg baseline \color{black}, where Cityscapes (C) and KITTI (K) serve as source domains and BDD100K (B) is the target. The yellow and blue bars indicate the results of local models using Cityscapes, KITTI, respectively. The green line gives the APs of the global model after fusing the client models. }
  \label{fig:TSNE}
\end{figure}
 
However, as shown in Figure~\ref{fig:TSNE}, we observe that in the early training rounds, the global model often underperforms the local models in representing target data.\footnote{\color{blue}This trend is obtained under the standard FedAvg baseline (without our proposed probabilistic or contrastive components).\color{black}} This observation challenges Moon’s assumption and motivates us to reverse the contrastive role—treating the global model as the negative sample instead:
\vspace{0.5em}
\begin{equation}
\label{equ:invmoon}
\mathcal{L}_{moon}' = -\log \left( \frac{\exp\left(\text{sim}(z, z_{\text{prev}})/ \tau\right)}{\exp\left(\text{sim}(z, z_{\text{glob}})/\tau\right) + \exp\left(\text{sim}(z, z_{\text{prev}})/\tau\right)} \right)
\end{equation}
\vspace{0.5em}

While local models may provide more reliable features in early stages, the global model gradually improves with aggregation and becomes more beneficial in later rounds. To reflect this evolving importance, we propose a \textit{dynamic model contrastive strategy} that interpolates between these two losses using a balancing factor $\lambda$:

\begin{equation}
\label{equ:Lcoin}
\begin{aligned}
\mathcal{L}_{\text{con}} 
&= - \left( \lambda \mathcal{L}_{\text{moon}} + (1 - \lambda) \mathcal{L}_{\text{moon}}' \right) \\
&= -\log \left( \frac{ \left( \exp\left( \text{sim}(z, z_{\text{glob}})/\tau \right) \right)^{1 - \lambda} 
                \cdot \left( \exp\left( \text{sim}(z, z_{\text{prev}})/\tau \right) \right)^{\lambda} }
         { \exp\left( \text{sim}(z, z_{\text{glob}})/\tau \right) 
         + \exp\left( \text{sim}(z, z_{\text{prev}})/\tau \right) } \right) \\
&= -\log \left( \frac{ \exp\left( \left( (1 - \lambda) \cdot \text{sim}(z, z_{\text{glob}}) + \lambda \cdot \text{sim}(z, z_{\text{prev}}) \right)/\tau \right) }
         { \exp\left( \text{sim}(z, z_{\text{glob}})/\tau \right) 
         + \exp\left( \text{sim}(z, z_{\text{prev}})/\tau \right) } \right).
\end{aligned}
\end{equation}

where \(\lambda \in [0, 1]\) balances the alignment between the global and previous local representations.

Combining the probabilistic localization loss $\mathcal{L}_{p}$ from the previous section with the contrastive objective $\mathcal{L}_{con}$, the total loss for each client is defined as:
\begin{equation}
\label{eq:L_client}
\mathcal{L}_{client} = \mathcal{L}_{p}+\mathcal{L}_{con}
\end{equation}
This formulation integrates the probabilistic localization loss to better capture uncertainty in bounding box regression, while dynamically balancing global and local representations through contrastive learning.

\subsubsection{Model Fusion Algorithm}
\label{sub:bakcbone_agg}

During FL, each client is trained using the Probabilistic Faster R-CNN described in Section~\ref{subsub:PFRCNN} and the dynamic model contrastive strategy introduced in Section~\ref{subsub:dynamic_moon}, as formulated in Eq.~\ref{eq:L_client}. At the end of each round, the server collects model parameters \{$M_{1}$, $M_{2}$, ..., $M_{N}$\} from $N$ clients. While standard FL aggregates the entire model, including both the backbone and detection head, our method only fuses the backbone due to architectural heterogeneity across clients. The local detection heads ($H_i$) will be retrained via knowledge distillation in Stage~3 (Sec. \ref{sub:KD}). 
\color{blue}The standard federated averaging scheme is adopted for server–client communication. At the beginning of each federated round \( r \), the server dispatches a shared initial model \( M^{r-1} \) to all \( N \) clients. Each client performs local training on its private data by minimizing the loss function $\mathcal{L}$ over mini-batches \( k \) for \( E \) iterations. After completing local updates, the resulting models are sent back to the server. The server then computes the average of these models to produce the updated global model \( M^r \). This model is subsequently distributed to all clients, and the same procedure is repeated for the remaining \( R-1 \) rounds.

% reviewer fix - <R2-M6> - waue 0922
\begin{comment} 
\begin{algorithm}[htbp]
\footnotesize
\caption{FedAvg}
\label{alg:fedavg}

\textbf{Input:} $N$ source domains $\{S_i\}_{i=1}^{N}$; target domain $T$; detectors $\{M^1, M^2, ..., M^N\}$; rounds $R$; local epochs $E$ \\
\textbf{Output:} final model $M^R$ \\
\textbf{Initialization:} Server initializes and sends $M^0$ to all $N$ clients

\begin{algorithmic}
\State \textbf{Server:}
\For{$r=1,...,R$}
    \For{$i=1,...,N$} \Comment{Local update}
        \State Adopt $M^{r-1}$ as initial model
        \State $M_i^r \gets \textproc{ClientUpdate}(i, M^{r-1})$
    \EndFor
    \State $M^r \gets \frac{1}{N} \sum_{i=1}^N M_i^r$
\EndFor
\State \Return $M^R$

\vspace{1mm}
\Statex \textbf{ClientUpdate($i$, $M^{global}$):}
\For{$j=1,...,E$}
    \State Sample mini-batch $\{(X^{s_i}_b, B^{s_i}_k, C^{s_i}_k)\}_{k=1}^{n_{s_i}}$
    \State Compute object detection loss $\mathcal{L}$
\EndFor
\State Update $M^{local}$ based on $\mathcal{L}$
\State \Return $M^{local}$
\end{algorithmic}
\end{algorithm}
\end{comment}
% reviewer fix - <R2-M6> - waue 0922

\color{blue}
Although more advanced model fusion strategies exist, such as FedMA~\cite{FedMA} and FedProx~\cite{li2020fedprox}, we adopt FedAvg as our default backbone fusion method due to its simplicity, effectiveness, and more consistent performance across datasets. Detailed comparisons and analysis are provided in the ablation study (Sec.~\ref{sub:ablation_model_fusion}).

\color{black}

\subsection{Stage 3: Cross-Client Knowledge Distillation}
\label{sub:KD}

As illustrated in Table~\ref{table:dataset_stats}, the source domains used in our experiments provide varying levels of category coverage. For instance, SIM10K contains only the "car" category, while KITTI includes five categories and Cityscapes covers eight. This inconsistency in categories leads to heterogeneous model architectures across domains, making it impractical to apply simple fusion algorithms that directly average the predictions from different detection heads. To tackle this issue, we propose the ICI algorithm, which explicitly matches and merges predictions from different sources at the category and instance levels.

\begin{algorithm}

\caption{ \textit{Inconsistent Categories Integration (ICI)}}
\label{alg:pseudo_proposals}
\textbf{Input:} pseudo-labels $\{\{p^i_j\}^{J_i}_{j=1}\}^N_{i=1}$ generated by the $N$ teacher heads $\{H^i\}_{i=1}^N$
\\
\textbf{Output:} union of integrated pseudo-labels $P$
\begin{algorithmic}

%\Require $L$: categories in target data
%\State $J_i$: proposals of the $i$-th client

\For{each $l \in L$}   \Comment{$L$: categories in target data}
    \State Initialize an empty set $C$
    \For{$i = 1$ to $N$}
        \If{$S_i$ contains label $l$}  \Comment{$S$: source data}
            \State $C \leftarrow C \cup \{s_i\}$   
            \Comment{$s_i$: index of source $i$}
        \EndIf
    \EndFor
    \If{$|C| \geq 1$}
        \For{$i = 1$ to $N$}
            \For{$j = 1$ to $J_i$} \Comment{$J$: number of proposals }
                %\State $t^{i}_j$: the bounding box of the \textit{j-th} region proposal of client $i$
                \State $\left(c^i_j, (b_x, b_y, b_w, b_h)^i_j\right) = p^i_j$ \Comment{$c$: class, $b$: bbox}
                \If{$\prod_{\substack{n=1, n \neq i}}^{N} \prod^{J_n}_{m=1} IoU(b^i_j, b^n_m) \neq 0$}
                %\State $d^i_j = \prod_{\substack{n=1, n \neq i}}^{N} \prod^{J_n}_{m=1} IoU(t^i_j, t^n_m)$
                %\Statex \Comment{$d$: IoU consistency index}
                    
                    \State $P_{l}\leftarrow P_{l} \cup \{p^i_j\}$% | d^i_j\neq0, \forall j \in J_i, \forall i \in N\}$ 
                    %\Statex \Comment{$p^i_j$: pseudo-labels generated by the $i$-th teacher}
                \EndIf
            \EndFor
        \EndFor

    \ElsIf{$|C| = 1$}
            \State $P_{l} \leftarrow \{p^{s^*}: s^* \in C\}$
    \EndIf
\EndFor
\Return $P = \{ P_l \mid l \in L \}$
\end{algorithmic}
\end{algorithm}

Algorithm~\ref{alg:pseudo_proposals} details our integration strategy based on spatial alignment and category consensus among the pseudo-labels. We first freeze the global backbone $M_g$ and use the local detection heads $H_i$ from different clients as multiple teachers to guide the training of a unified student head $H_g$. For each target category $l \in L$, the algorithm first selects teacher heads whose source datasets contain this category. When multiple heads share the category, ICI performs spatial alignment by evaluating the Intersection-over-Union (IoU) between all proposals across teachers. Proposals that have non-zero IoU with predictions from other heads are retained, while isolated or conflicting ones are discarded. Specifically, for a proposal with bounding box $b_j^i$, it is selected into the final pseudo-label pool $P_l$ if:

\[
\prod_{\substack{n=1, n\neq i}}^{N}\prod^{J_n}_{m=1}IoU(b^i_j,b^n_m) \neq 0.
\]

This ensures that only consistent regions across domains are used for supervision, while proposals from a single teacher are directly included when no cross-source agreement is available. The resulting pseudo-label pool $P = \{P_l \,|\, l \in L\}$ is then used to compute the distillation loss $\mathcal{L}_{kd}$, which has the same form as $\mathcal{L}_{p}$ but uses ICI-generated labels as targets.

% --------------------------------------------------------------------------------------

\section{Experiments}
\label{sec:experiments}

In this section, we conduct extensive experiments to evaluate the effectiveness of the proposed \textit{HeroCrystal}. We first describe the experimental setup, including datasets, evaluation metrics, and implementation details in Sec. \ref{subsec:setup}. Then, we benchmark \textit{HeroCrystal} against a variety of state-of-the-art (SOTA) methods under diverse domain adaptation scenarios in Sec. \ref{subsec:results}. Finally, we present ablation studies to access the contribution of each components in Sec. \ref{subsec:ablation}.

\subsection{Experimental Setup}
\label{subsec:setup}

This section describes the experimental setup used to evaluate our proposed framework. We begin by introducing the datasets involved in multi-source domain adaptation (Sec. \ref{sub:dataset}), followed by the evaluation metrics and performance protocol (Sec. \ref{sub:evaluation-metrics}). Implementation details regarding model configuration and training schedules are presented in Sec. \ref{sub:implementation-details}. Finally, Sec \ref{subsec:baselines} summarizes the baseline and ablation methods used for comparison.

\subsubsection{Datasets} 
\label{sub:dataset}
Our experiments are benchmarked on five widely-used public datasets for autonomous driving scenarios: \textbf{Cityscapes (C)} \cite{cordts2016cityscapes}, its synthetic foggy variant \textbf{Foggy Cityscapes (F)} \cite{Sakaridis2018Foggy}, \textbf{KITTI (K)} \cite{Geiger2012KITTI}, the synthetic dataset \textbf{SIM10K (S)} \cite{Johnson-Roberson2017SIM10K}, and the large-scale, diverse \textbf{BDD100K (B)} \cite{Yu2020BDD100K}. These datasets exhibit significant domain shifts in terms of weather conditions, camera viewpoints, and scene complexity, making them ideal for evaluating domain adaptation performance.

A key challenge in multi-source training is the category inconsistency across datasets. For instance, the category for humans is labeled as `person` in some datasets and `pedestrian` in others. To ensure a consistent evaluation, we unify the labels into eight common categories: \textit{bicycle, bus, car, motorcycle, person, rider, train,} and \textit{truck}. A summary of the datasets and their corresponding object categories is provided in Table \ref{table:dataset_stats}.

To simulate realistic multi-source domain adaptation, we construct two integrated experimental settings. In the \textbf{\textit{CK$\rightarrow$B}} setting, we use \textit{Cityscapes (C)} and \textit{KITTI (K)} as source domains and \textit{BDD100K (B)} as the target domain. In the \textit{\textbf{SKF$\rightarrow$C}} setting, we combine \textit{SIM10K (S)}, \textit{KITTI (K)}, and \textit{Foggy Cityscapes (F)} as source domains, with \textit{Cityscapes (C)} as the target. These settings cover a broad range of visual shifts and label distributions, making them suitable for evaluating generalization under category and domain mismatch.

\begin{comment}
    
%% this is table 4 (replace by table 3 )
\begin{table}[htbp]
    \centering
    \caption{Summary of datasets used in our experiments. We list the number of training images, the domain type (Source/Target), and the available object categories after unification.}
    \label{table:dataset_summary}
    \resizebox{.8\columnwidth}{!}{%
    \begin{tabular}{c c c l}
        \toprule
        \textbf{Dataset} & \textbf{\# Images} & \textbf{Domain Role} & \makecell{\textbf{Unified Categories} \\ \textbf{(Count)}}
\\
        \midrule
        \makecell[c]{ Cityscapes \\(C) }& 2,975 & Source / Target & \makecell[l]{bicycle, bus, car,\\ motorcycle, person, \\rider, train, truck (8)} \\
         \midrule
        \makecell[c]{Foggy Cityscapes\\ (F)} & 2,975 & Source &  \makecell[l]{bicycle, bus, car,\\ motorcycle, person, \\rider, train, truck (8)} \\
         \midrule
        \makecell[c]{KITTI \\(K)} & 7,481 & Source & \makecell[l]{car, person, rider,\\ train, truck (5)} \\
         \midrule
        \makecell[c]{SIM10K \\(S)} & 10,000 & Source & car (1) \\
         \midrule
        \makecell[c]{BDD100K \\(B)} & 7,000 & Target &  \makecell[l]{bicycle, bus, car,\\ motorcycle, person, \\rider, train, truck (8)} \\
        \bottomrule
    \end{tabular}%
    }
\end{table}
\end{comment}

\subsubsection{Evaluation Metrics} 
\label{sub:evaluation-metrics}

We follow the standard evaluation protocol for object detection, where the performance is measured using average precision (AP) for each traffic category (e.g., \textit{car}, \textit{train}, \textit{truck}) and mean average precision (mAP) across all categories. For a given category \( c \), the AP is computed as the area under the precision--recall curve, defined as \( \text{AP}_c = \int_0^1 p_c(r) \, dr \), where \( p_c(r) \) denotes the precision as a function of recall. We report per-category AP values, such as \( \text{AP}_{\textit{car}} \), \( \text{AP}_{\textit{truck}} \), and \( \text{AP}_{\textit{train}} \). The overall performance is then summarized by the mean of AP values across all \( C \) categories, i.e., \( \text{mAP}_{\textit{all}} = \frac{1}{C} \sum_{c=1}^C \text{AP}_c \). In our experiments, the Intersection over Union (IoU) threshold is fixed at 0.5, and all reported results are based on this mAP@50 metric.

\begin{comment}
    
We follow the standard evaluation protocol for object detection. The performance is evaluated using Average Precision (AP) for each traffic category (e.g., \textit{car}, \textit{train}, \textit{truck}) and the mean Average Precision (mAP) across all categories. 

For a given category \( c \), the AP is defined as the area under the precision–recall curve:

\begin{equation}
\text{AP}_c = \int_0^1 p_c(r) \, dr
\end{equation}

where \( p_c(r) \) denotes the precision as a function of recall for category \( c \).  
To evaluate per-category performance, we report \( \text{AP}_{\textit{car}} \), \( \text{AP}_{\textit{truck}} \), \( \text{AP}_{\textit{train}} \), etc.  
The overall performance is measured by the mean of AP values across all \( C \) categories:

\begin{equation}
\text{mAP}_{\textit{all}} = \frac{1}{C} \sum_{c=1}^C \text{AP}_c
\end{equation}

In our experiments, the Intersection over Union (IoU) threshold is fixed at 0.5. All reported results are based on this mAP@50 metric.
\end{comment}

\subsubsection{Implementation Details} 
\label{sub:implementation-details}
In the \textit{Generated} stage, we set the maximum training steps to 1000 and the batch size to 2. The training process uses a fixed learning rate of \(1 \times 10^{-6}\), and the weighting coefficient \(\alpha\) for the total loss is set to 1 to ensure style consistency while preserving both diversity and semantic fidelity in the generated images. During inference, we configure the number of inference steps to 100 and set the guidance scale to 7.5. The long-tailed subset used in the \textit{HeroCrystal-rare} setting includes the following object classes: \textit{bus}, \textit{motorcycle}, \textit{truck}, and \textit{train}, which have significantly fewer instances across domains.

In the \textit{Federated} and \textit{Distilled} stages, all experiments are implemented using the Detectron2~\cite{wu2019detectron2} framework with a VGG16~\cite{simonyan2014very} backbone pre-trained on ImageNet. For the FL process, we set the total number of communication rounds to \(R = 3\), with 4000 local iterations per client in each round. The model is trained using stochastic gradient descent (SGD) with a momentum of 0.9 and a learning rate of 0.001. For the contrastive learning parameters in Equation~\ref{equ:Lcoin}, we set the temperature \(\tau = 0.5\) and define the dynamic weighting factor as \(\alpha = \frac{1}{r}\), where \(r\) denotes the current round.

\subsubsection{Compared Methods}
\label{subsec:baselines}

We compare our framework with a comprehensive set of methods spanning different categories to demonstrate its unique advantages.

% waue revise 20251001
\color{blue}
\begin{itemize}
    \item \textbf{Source-only}: Lower-bound baseline where classic detectors (Faster R-CNN~\cite{faster-RCNN}, FCOS~\cite{fcos}, Def DETR~\cite{deformable-DETR}) are trained only on source domain(s) and tested directly on the target domain, revealing the extent of the cross-domain gap. 
    
    \item \textbf{Domain Adaptation}: 
Single-source DA — Domain adaptation from a single source domain, including methods such as DA-faster~\cite{chen2018domain}, SW~\cite{saito2019strong}, PET~\cite{liu2023periodically}, SIGMA~\cite{li2022sigma}, TDD~\cite{he2022cross}, OADA~\cite{yoo2022oada}, SFA~\cite{wang2021exploring}, AQT~\cite{huang2022aqt}, and MRT~\cite{zhao2023masked}. These approaches illustrate the limitation of not leveraging multiple source domains. 
Multi-source DA — Trained with data aggregated from multiple source domains in a centralized manner, such as PMT~\cite{PMT2024} and MTK~\cite{zhang2022multi}. While such models often achieve competitive performance, they \textbf{violate data privacy}, a core constraint respected by our method.

    \item \textbf{Standard FL}: Privacy-preserving federated learning frameworks, including FedAvg~\cite{FedAvg}, FedMA~\cite{FedMA}. However, their primary drawback is the inability to handle clients with \textbf{inconsistent output categories}, limiting applicability in realistic multi-domain scenarios.
    
    \item \textbf{Multi-class FL}: Federated learning methods with support for category-inconsistent clients or enhanced aggregation, including Moon+ICI~\cite{moon}, FedCoin~\cite{FedCoin}, FedCoin\_FedProx, with results evaluated across diverse classes. These are strong competitors leveraging federated representation learning.
    
    \item \textbf{HeroCrystal (Ours)}: Our proposed method, HeroCrystal-all and HeroCrystal-rare, builds upon strong federated learning baselines and introduces a novel federated data augmentation approach to enhance model generalization and performance on the target domain.
    
    \item \textbf{Oracle}: An upper-bound reference in which the model is trained using supervised learning with centralized labeled target domain data (via Faster R-CNN), used in both Table 4 and Table 5 to assess the practical gap. This does not consider privacy-preserving constraints.
\end{itemize}
 \color{black}

\subsection{Main Results on Cross-Domain Adaptation}
\label{subsec:results}

To assess our method under diverse domain shifts, we design two settings: \textit{CK$\rightarrow$B} for adapting from standard to diverse scenes (Sec. \ref{subsec:results_ck2b}), and \textit{SKF$\rightarrow$C} for adapting from mixed sources to clear weather (Sec. \ref{subsec:results_skf2c}). These settings reflect real-world challenges in category and domain mismatches.

%We evaluate all methods on two challenging adaptation scenarios: (1) \textbf{CK$\rightarrow$B}: from Cityscapes and KITTI to BDD100K, and (2) \textbf{SKF$\rightarrow$C}: from SIM10K, KITTI, and Foggy Cityscapes to Cityscapes.

% 20251002 waue revised
% table 4
\begin{table*}[htbp]
\centering
\caption{Quantitative Results for Multi-Source Domain Adaptation (\textit{CK$\rightarrow$B}).}
\label{table:results_ck_b}

\resizebox{\textwidth}{!}{%
\begin{threeparttable}
\begin{tabular}{c l c c ccc cccccccc}
\toprule
\makecell{\textbf{Type}} & \textbf{Method} & \color{blue}\textbf{Detector} \color{black}& \makecell{\color{blue}\textbf{Source}\color{black}} & \multicolumn{3}{c}{\textbf{Properties}} & \multicolumn{8}{c}{\textbf{AP (\%) on Target BDD100K ($\uparrow$)}} \\
\cmidrule(lr){5-7} \cmidrule(lr){8-15}
& & & & $P_r$ & $I_m$ & $O_m$ & car & truck & rider & person & motor & bicycle & bus & \textbf{mAP} \\
\midrule
\multirow{3}{*}{\makecell{Source\\-only}} 
    & FRCNN \cite{faster-RCNN} & FRCNN & C &  & \checkmark & \checkmark & 44.7 & 17.4 & 22.1 & 26.9 & 17.1 & 18.8 & 16.7 & 23.4 \\
    & \color{blue}FCOS \cite{fcos} & FCOS & C &   &   & \checkmark & 54.5 & 17.2 & 24.8 & 38.6 & 15.0 & 18.3 & 16.3 & 26.4 \\
    & \color{blue}Def DETR \cite{deformable-DETR} & Def DETR & C &   &   & \checkmark & 55.2 & 15.7 & 26.7 & 38.9 & 10.8 & 16.2 & 19.7 & 26.2 \\
\midrule
\multirow{7}{*}{\makecell{Domain\\Adaption}}
    & DA-faster \cite{chen2018domain} & FRCNN & C &  &  & \checkmark & 44.6 & 14.3 & 26.5 & 29.4 & 15.8 & 20.6 & 16.8 & 24.0 \\
    & SW \cite{saito2019strong} & FRCNN & C &  &  & \checkmark & 45.7 & 15.2 & 29.5 & 30.2 & 17.1 & 21.2 & 18.4 & 25.3 \\
    & \color{blue}PET \cite{liu2023periodically} & FRCNN & C & & & \checkmark & 62.4&19.3&34.5&42.6&17.0&26.3&16.9&31.3 \\
    & \color{blue}SIGMA \cite{li2022sigma} & FCOS & C &   &   & \checkmark & \textbf{64.1} & 20.2 & 29.6 & \textbf{46.9} & 17.9 & 26.3 & 23.6 & 32.7 \\
    & \color{blue}SFA \cite{wang2021exploring} & Def DETR & C &   &   & \checkmark & 57.5 & 19.1 & 27.6 & 40.2 & 15.4 & 19.2 & 23.4 & 28.9 \\
    & \color{blue}AQT \cite{huang2022aqt} & Def DETR & C &   &   & \checkmark & 58.4 & 17.3 & 33.0 & 38.2 & 16.9 & 23.5 & 18.4 & 29.4 \\
    & PMT \cite{PMT2024} & FRCNN & C,K &  & \checkmark &  & 58.7 & -- & -- & -- & -- & -- & -- & -- \\
\midrule
\multirow{2}{*}{\makecell{Standard\\FL}}
    & FedAvg \cite{FedAvg} & FRCNN & C,K & \checkmark & \checkmark &  & 44.1 & -- & -- & -- & -- & -- & -- & -- \\
    & FedMA \cite{FedMA} & FRCNN & C,K & \checkmark & \checkmark &  & 46.0 & -- & -- & -- & -- & -- & -- & -- \\
\midrule
\multirow{4}{*}{\makecell{Multi-class\\FL}}
    & Moon \cite{moon}+ICI & FRCNN & C,K & \checkmark & \checkmark & \checkmark & 50.0 & 21.6 & 32.6 & 30.1 & 15.2 & 27.3 & 22.1 & 28.4 \\
    & FedCoin \cite{FedCoin} & FRCNN & C,K & \checkmark & \checkmark & \checkmark & 52.4 & 22.2 & 35.5 & 35.5 & 19.1 & 29.1 & 25.1 & 31.3 \\
  %  & \color{red}FedCoin_FedProx & FRCNN & C,K & \checkmark & \checkmark & \checkmark & 37.3 & 16.2 & 21.0 & 36.4 & 17.9 & 28.3 & 21.9 & 22.4 \\
    &  \textbf{HeroCrystal-all}\textsuperscript{\textdagger} & FRCNN & C,K & \checkmark & \checkmark & \checkmark & 49.7 & 27.5 & 35.4 & 36.2 & 21.1 & 29.9 & 27.5 & 32.5 \\
    & \textbf{HeroCrystal-rare}\textsuperscript{\textdaggerdbl} & FRCNN & C,K & \checkmark & \checkmark & \checkmark & 52.8 & \textbf{27.7} & \textbf{36.1} & 37.2 & \textbf{21.6} & \textbf{31.2} & \textbf{27.7} & \textbf{33.5} \\
\midrule
\multirow{1}{*}{Oracle} 
    & Supervised & FRCNN & B &  &  & \checkmark & 53.9 & 46.3 & 33.2 & 35.3 & 25.6 & 29.3 & 46.7 & 38.6 \\
\bottomrule
\end{tabular}
\begin{tablenotes}
\item[*] $P_r$ denotes privacy-preserving, $I_m$ represents multi-source, and $O_m$ stands for multi-class.
\item[*] Source Dataset, C = Cityscapes, K = KITTI, B = BDD100K.
\item[*] Best results among privacy-preserving methods are in \textbf{bold}.
\item[*] \color{blue} \textbf{HeroCrystal-all}\textsuperscript{\textdagger} uniformly augments all categories, while \textbf{HeroCrystal-rare}\textsuperscript{\textdaggerdbl} selectively augments long-tailed classes to improve rare-category performance.
\item[*] \color{red} Results of non-FRCNN baselines (e.g., FCOS, Deformable DETR) are cited from the original papers and used as reference baselines.
\end{tablenotes}
\end{threeparttable}
}
\end{table*}

\subsubsection{From Standard to Diverse Scenes (CK$\rightarrow$B)}
\label{subsec:results_ck2b}

\color{blue}
Table~\ref{table:results_ck_b} presents the quantitative results for the \textit{CK$\rightarrow$B} scenario. For the single-source DA setting, HeroCrystal consistently outperforms several privacy-violating domain adaptation baselines such as DA-faster~\cite{chen2018domain} and SW~\cite{saito2019strong}, while maintaining strong privacy guarantees. Recent advanced DA methods, including PET~\cite{liu2023periodically}, SIGMA~\cite{li2022sigma}, SFA~\cite{wang2021exploring}, and AQT~\cite{huang2022aqt}, achieve higher performance with complex regularization or attention designs but rely on centralized supervision and lack privacy preservation. Among these, SIGMA~\cite{li2022sigma} achieves the highest accuracy for the \textit{car} category (64.1\%), although it remains a single-source method without FL mechanisms. The multi-source DA method PMT~\cite{PMT2024} also achieves competitive high accuracy for the \textit{car} category (58.7\%), but it supports only single-category adaptation, reports results for one class, and disregards privacy constraints. In contrast, our proposed HeroCrystal framework simultaneously handles diverse categories across distributed clients, preserves source data privacy, and supports multi-source domain adaptation, making it more scalable and practical for real-world applications.
\color{black}

In addition, compared to methods that consider privacy, standard FL baselines such as FedAvg (44.1\% mAP) and FedMA (46.0\% mAP) fail to effectively address label space heterogeneity. In contrast, HeroCrystal achieves significantly better performance, reaching 49.7\% (HeroCrystal-all) and 52.8\% mAP (HeroCrystal-rare), respectively. Compared to other contrastive FL methods (such as FedCoin), our proposed HeroCrystal-all outperforms it in nearly all categories except \textit{car}, while  HeroCrystal-rare surpasses it across all categories and achieves a mAP of 33.4\%, with a margin of +2.1\%. These results highlight the effectiveness of our approach under federated settings.

We further compare two variants of HeroCrystal: HeroCrystal-all, which uniformly augments all categories, and HeroCrystal-rare, which selectively generates synthetic samples for long-tailed classes. While uniform augmentation may introduce redundancy or degrade performance in dominant categories such as \textit{car}, the rare-category-focused strategy achieves higher overall stability and accuracy, culminating in a new state-of-the-art 33.4\% mAP among privacy-preserving methods. This highlights the effectiveness of targeted augmentation in mitigating class imbalance and improving cross-domain transferability without sacrificing general performance---an insight further supported by the detailed ablation results in Section~\ref{sub:generate_single_ablation}.

\begin{table*}[htbp]
\centering
\caption{Quantitative Results for Multi-Source Domain Adaptation (\textit{SKF$\rightarrow$C}).}
\label{table:results_skf_c}

\resizebox{\textwidth}{!}{%
\begin{threeparttable}
\begin{tabular}{c l c c ccc cccccccc}
\toprule
\makecell{\textbf{Type}} & \textbf{Method} & \color{blue} \textbf{Detector} & \makecell{\color{blue} \textbf{Source}} & \multicolumn{3}{c}{\textbf{Properties}} & \multicolumn{8}{c}{\textbf{AP (\%) on Target Cityscapes ($\uparrow$)}} \\
\cmidrule(lr){5-7} \cmidrule(lr){8-15}
& & & & $P_r$ & $I_m$ & $O_m$ & car & truck & rider & person & motor & bicycle & bus & \textbf{mAP} \\
\midrule
\multirow{1}{*}{\makecell{Source-only}}
    & FRCNN \cite{faster-RCNN} & FRCNN & F & & \checkmark & & 44.2 & -- & -- & -- & -- & -- & -- & -- \\
\midrule
\multirow{9}{*}{\makecell{Domain\\Adaption}}
    & DA-faster \cite{chen2018domain} & FRCNN & F & & & & 50.7 & -- & -- & -- & -- & -- & -- & -- \\
    & SW \cite{saito2019strong} & FRCNN & F & & &  & 51.8 & -- & -- & -- & -- & -- & -- & -- \\
    & \color{blue}PET \cite{liu2023periodically} & FRCNN & S & & &  & 57.8 & -- & -- & -- & -- & -- & -- & -- \\
    & \color{blue}SIGMA \cite{li2022sigma} & FCOS & S &  &  & & 53.7 & -- & -- & -- & -- & -- & -- & --  \\
    & \color{blue}TDD \cite{he2022cross} & FRCNN & S &  &  &  & 53.4 & -- & -- & -- & -- & -- & -- & --  \\
    & \color{blue}OADA \cite{yoo2022oada} & FCOS & S &  &  &  & 56.6 & -- & -- & -- & -- & -- & -- & --  \\
    & \color{blue}AQT \cite{huang2022aqt} & Def DETR & S &  &  &  & 53.4 & -- & -- & -- & -- & -- & -- & -- \\
    & \color{blue}MRT \cite{zhao2023masked} & Def DETR & S &  &  &  & 62.0 & -- & -- & -- & -- & -- & -- & --  \\
    & \color{blue}MTK \cite{zhang2022multi} & FRCNN & S,K,F & & \checkmark & & 52.9 & -- & -- & -- & -- & -- & -- & -- \\
\midrule
\multirow{2}{*}{\makecell{Standard\\FL}}
    & FedAvg \cite{FedAvg} & FRCNN & S,K,F & \checkmark & \checkmark & & 51.1 & -- & -- & -- & -- & -- & -- & -- \\
    & FedMA \cite{FedMA} & FRCNN & S,K,F & \checkmark & \checkmark & & 49.9 & -- & -- & -- & -- & -- & -- & -- \\
\midrule
\multirow{4}{*}{\makecell{Multi-class\\FL}}
    & Moon+ICI \cite{moon} & FRCNN & S,K,F & \checkmark & \checkmark & \checkmark & 60.7 & 31.4 & 43.6 & 38.2 & 28.7 & 37.1 & 47.0 & 41.0 \\
    & FedCoin \cite{FedCoin} & FRCNN & S,K,F & \checkmark & \checkmark & \checkmark & 62.7 & 32.0 & \textbf{45.3} & 37.3 & 30.4 & 39.5 & 50.8 & 42.9 \\
    & \textbf{HeroCrystal-all}\textsuperscript{\textdagger} & FRCNN & S,K,F & \checkmark & \checkmark & \checkmark & 65.2 & \textbf{40.2} & 42.7 & \textbf{40.2} & 32.0 & 40.8 & 58.3 & \textbf{45.6} \\
    & \textbf{HeroCrystal-rare}\textsuperscript{\textdaggerdbl}  & FRCNN & S,K,F & \checkmark & \checkmark & \checkmark & \textbf{65.7} & 36.1 & 38.4 & 39.7 & \textbf{32.2} & \textbf{43.3} & \textbf{63.6} & \textbf{45.6} \\
\midrule
\multirow{1}{*}{Oracle}
    & Supervised & FRCNN & C & & & \checkmark & 65.9 & 31.8 & 50.1 & 49.3 & 30.8 & 37.5 & 51.2 & 45.2 \\
\bottomrule
\end{tabular}

\begin{tablenotes}
\item[*] $P_r$ denotes privacy-preserving, $I_m$ represents multi-source, and $O_m$ stands for multi-class.
\item[*] Source Dataset, S = Sim10K, K = KITTI, F = Foggy Cityscapes, C = Cityscapes.
\item[*] Best results among privacy-preserving methods are in \textbf{bold}.
\item[*] \color{blue} \textbf{HeroCrystal-all}\textsuperscript{\textdagger} uniformly augments all categories, while \textbf{HeroCrystal-rare}\textsuperscript{\textdaggerdbl} selectively augments long-tailed classes to improve rare-category performance.
\item[*] \color{red} Results of non-FRCNN baselines (e.g., FCOS, Deformable DETR) are cited from the original papers and used as reference baselines.
\end{tablenotes}
\end{threeparttable}
}
\end{table*}

\subsubsection{From Mixed Sources to Clear Weather (SKF$\rightarrow$C)} 
\label{subsec:results_skf2c}

\color{blue}

To further validate the robustness of HeroCrystal, we conduct experiments on the \textit{SKF$\rightarrow$C} scenario, where the sources are a mix of synthetic, real-world, and adverse-weather data, as shown in Table~\ref{table:results_skf_c}. Among domain adaptation methods, recently proposed approaches---such as PET~\cite{liu2023periodically}, SIGMA~\cite{li2022sigma}, TDD~\cite{he2022cross}, OADA~\cite{yoo2022oada}, AQT~\cite{huang2022aqt}, and MRT~\cite{zhao2023masked}---are included for a more comprehensive comparison. Notably, MRT achieves the highest car AP (62.0\%) among these approaches; however, the absence of privacy preservation restricts their applicability in federated settings. These SOTA methods leverage various strategies such as periodic training, mask-based transfer, or multi-level alignment to boost adaptation, yet still lag behind multi-class FL methods in overall mAP. Meanwhile, earlier DA baselines (DA-faster~\cite{chen2018domain}, SW~\cite{saito2019strong}) and the multi-source variant MTK~\cite{zhang2022multi} also fall short in this mixed-source setting.

\color{black}
Among multi-class FL methods, our proposed \textbf{HeroCrystal} consistently outperform prior work in terms of mAP, with the only exception being the \textit{rider} category where FedCoin \cite{FedCoin} performs slightly better. However, unlike the more stable trend observed in \textit{CK$\rightarrow$B}, the performance advantage between HeroCrystal-all and HeroCrystal-rare varies across categories—for instance, \textit{truck} and \textit{person} benefit more from augmenting all categories. This inconsistency may be attributed to the larger number of clients and the more imbalanced category distribution in the \textit{SKF$\rightarrow$C} setup, introducing greater variability in category-wise across domains. Notably, for certain categories such as \textit{truck} and \textit{person}, the federated multi-source augmentation and class-selective strategies of HeroCrystal enable superior generalization compared to Oracle, which may not fully exploit the diverse source data and augmentation potential arising from federated training.

\subsection{Ablation Study}
\label{subsec:ablation}

To evaluate the effectiveness of each component in our proposed framework, we conduct a comprehensive ablation study covering three key aspects. First, Sec. ~\ref{sub:ablation_loss} analyzes the impact of individual loss terms on model performance. Second, Sec. ~\ref{sub:generate_single_ablation} investigates the role of target-aware generation under a single-category setting. Third, Sec. ~\ref{sub:ablation_model_fusion} compares different model fusion strategies used during the federated stages. Finally, in Sec.\ref{sub:FN_reduce_ablation}, we evaluate the impact of probabilistic modeling on the distribution of false negatives and false positives in pseudo-labels.

\subsubsection{Effect of Model Components}
\label{sub:ablation_loss}
To examine the effectiveness of individual components in our framework, we conduct a detailed ablation study on four key loss functions in different stages: $\mathcal{L}_{p}$ (Sec. \ref{subsub:PFRCNN}), $\mathcal{L}_{con}$ (Sec. \ref{subsub:dynamic_moon}), $\mathcal{L}_{kd}$ (Sec. \ref{sub:KD}), and $\mathcal{L}_{gen}$ (Sec. \ref{sub:augmented}). When used $\mathcal{L}_p$ alone under the single-source setting, it already significantly improves detection performance (e.g., \textit{car} improves from 44.11\% to 50.75\%), confirming the effectiveness of incorporating with localization uncertainty. For multi-source domain adaptation, due to the inconsistency of categories among clients, it is necessary to combine the ICI algorithm ($\mathcal{L}_{kd}$) with the probabilistic loss ($\mathcal{L}_{p}$). The resulting performance is comparable to using $\mathcal{L}_{p}$ alone, indicating that the ICI algorithm can effectively integrate different categories without degrading accuracy. Furthermore, adopting $\mathcal{L}_{con}$ during client training improves most categories except \textit{truck} and \textit{bicycle}.

\begin{table*} 
\centering
 \caption{Ablation study on \textit{CK$\rightarrow$B} showing the effect of each component. }
 \label{table:ablation}
  \begin{threeparttable}
  \resizebox{\textwidth}{!}{%
  \begin{tabular}{c|cccc|cccccccc}
   \toprule
   source &$ \mathcal{L}_{p}$ & $\mathcal{L}_{kd}$ & $\mathcal{L}_{con}$ & $\mathcal{L}_{gen}$  &
    car & truck & rider & person & motor & bicycle & bus &  mAP  \\
   \midrule
\multirow{2}{*}{single} 
&   &   &   && 44.11  & - & - & - & - & - & - & - \\	
&  \checkmark &   &   & &50.75 & - & - & - & - & - & - & - \\	
   \midrule
\multirow{4}{*}{multi} 
&  \checkmark &  \checkmark &   & &50.50 & 24.21 & 32.64 & 31.87 & 16.45 & 29.62 & 24.70 & 29.46\\	
&  \checkmark &  \checkmark &\checkmark & &52.39 & 22.22 & 35.51 & 35.52 & 19.07 & 29.10 & 25.09 &31.27\\	
& \checkmark & \checkmark & \checkmark & $all$ & 49.70 & 27.50 & 35.40 & 36.20 & 21.10 & 29.90 & 27.50 & 32.47 \\
& \checkmark & \checkmark & \checkmark & $rare$  &  \textbf{52.80} &  \textbf{27.70} &  \textbf{36.10} &  \textbf{37.20} &  \textbf{21.60} &  \textbf{31.20} &  \textbf{27.70} &  \textbf{33.47} \\
   \bottomrule
  \end{tabular}
  }
 \end{threeparttable}
\end{table*}

To address the low accuracy of long-tailed categories, we introduce a target-aware generation module guided by $\mathcal{L}_{\text{gen}}$. When augmenting all categories, performance on dominant classes (e.g., \textit{car}) slightly declines, likely because the synthetic images frequently featuring large vehicles (Figure~\ref{fig:bddgen_example}(a)), which do not effectively enhance the model’s ability to detect smaller objects. To mitigate this, we restrict augmentation to rare classes (e.g., \textit{motorcycle}, \textit{truck}, \textit{bus}, \textit{train}), leading to notable gains, especially for long-tailed categories, such as \textit{truck} (+5.4\%) and \textit{motor} (+2.5\%). These results confirm that selectively augmenting rare classes is more effective than indiscriminate augmentation, highlighting the advantage of our target-aware generation module in enhancing long-tail generalization.

\begin{figure}[htb]
\centering

   \includegraphics[width=\linewidth]{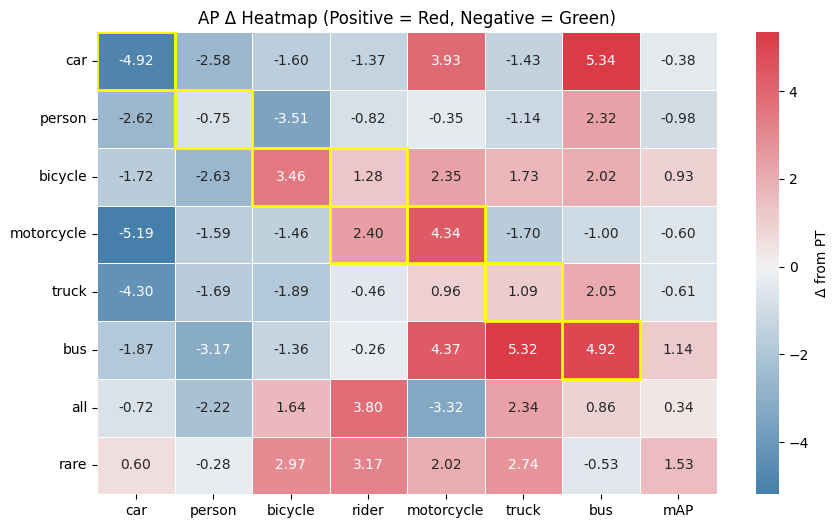}

  \caption{AP differences from the PT\cite{chen2022PT} baseline under category-specific generation. Red/blue cells indicate gains/drops. Each row is a generated category, each column an evaluated class. Yellow frames highlight self-impact or related improvements.}
  \label{fig:heatmap}
\end{figure}

\color{black}
\subsubsection{Effectiveness of Target-Aware Generation}
\label{sub:generate_single_ablation}

\noindent
\textbf{Target-Aware Generation for Single-Category Adaptation.}

To evaluate the effectiveness of category-specific image generation, we conduct an ablation study using the PT method \cite{chen2022PT}, a single-source domain adaptive object detection baseline chosen for its faster training efficiency over FL. In this experiment, we augment the training data by adding synthetic images for only one category at a time and observe how this affects detection performance across all categories. As shown in Figure~\ref{fig:heatmap}, generating images for long-tailed categories leads to noticeable improvements, as indicated by the predominantly red-colored cells on the right part. In contrast, generating data for well-performing common categories such as \textit{car} and \textit{person} often yields negative gains, reflected by the blue-colored cells on the left part.

\begin{figure}[htb]
\centering
\subfloat[]{
   \includegraphics[width=.4\linewidth]{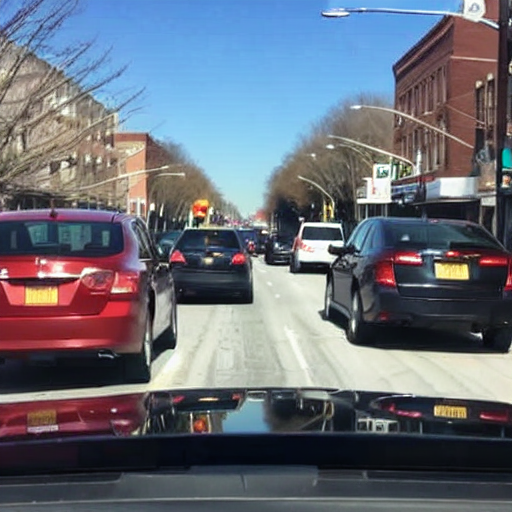}}
   \subfloat[]{
   \includegraphics[width=.4\linewidth]{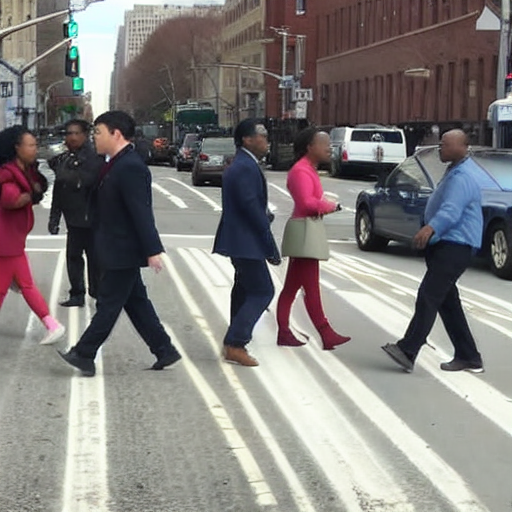}
   }

  \caption{Qualitative examples of generated images in BDD100K. (a)Person. (b)Car.}
  \label{fig:bddgen_example}
\end{figure}

The yellow-framed cells in the heatmap indicate the self-impact of generating a category on its own performance (e.g., generating \textit{bus} improves \textit{bus}). Since \textit{rider} typically appears alongside \textit{motorcycle} or \textit{bicycle}, we omit direct generation of \textit{rider} and instead consider the augmentation of these co-occurring categories as indirect enhancements for it, as reflected in the highlighted cells. As shown, most yellow-framed categories generally show positive gains except for \textit{person} and \textit{car}, indicating that that category-specific generation is generally beneficial. The performance drop for \textit{person} is due to artifacts or lack of facial clarity in the generated images, as illustrated in Figure~\ref{fig:bddgen_example}(a). For \textit{car}, a dominant category, the synthetic images often depict large vehicles (Figure~\ref{fig:bddgen_example}(b)), which fails to address the real challenge of detecting small, distant cars in BDD100K and may reduce performance on small-object detection due to distribution mismatch. However, in the multi-source setting, the ICI algorithm mitigates the performance degradation of \textit{person} and \textit{car} by aggregating knowledge from multiple sources like KITTI.

We compare different generation strategies and find that augmenting \textit{All} categories simultaneously improves overall performance but causes noticeable drops in some key classes (e.g., \textit{person}: $-2.22\%$, \textit{motorcycle}: $-3.32\%$). In contrast, generating only long-tail categories (\textit{rare}) yields mostly positive gains with minimal losses (e.g., \textit{person}: $-0.28\%$, \textit{bus}: $-0.53\%$), indicating a better balance between improvement and stability. Based on these findings, we adopt both long-tail-only and full-category generation in our final multi-camera FL setup to achieve broader coverage while retaining effectiveness.

\color{blue}
\noindent
\textbf{Quantitative Evaluation of Target-Aware Generation.}

\begin{table}[htbp]
\centering
\caption{\color{blue}Quantitative comparison of generation quality among different generation strategies on the Cityscapes dataset.}
\label{tab:fid}
\begin{threeparttable}
\resizebox{\textwidth}{!}{%
\begin{tabular}{lccc|c}
\toprule
\multirow{2}{*}{}& 
\multicolumn{3}{c|}{\textbf{Generation Methods}} &
\multirow{2}{*}{\textbf{FID ($\downarrow$)}}  \\
\cmidrule(lr){2-4}\cmidrule(lr){5-5}
 & \textbf{Layout Diffusion } & \textbf{Style Transfer } & \textbf{Target-Aware} 
 & \\
\midrule
Layout Diffusion \cite{zheng2023layoutdiffusion}& \checkmark &  &  & 283.4 \\
Two-Stage \cite{zheng2023layoutdiffusion} + \cite{johnson2016FastNeuralStyleTransferNetwork}& \checkmark & \checkmark &  & 263.1  \\
\textbf{Ours} &  & \checkmark & \checkmark & \textbf{192.6*} \\
\bottomrule
\end{tabular}
} % resize
\begin{tablenotes}
\scriptsize
\item[*] The reported value for \textbf{Ours} is the average FID across all object categories; \\Detailed per-class FID scores are presented in Table~\ref{tab:fid_classes}.
\end{tablenotes}
\end{threeparttable}
\end{table}

\begin{table}[htbp]
\centering
\caption{ \color{blue}Quantitative evaluation of generated images across different object categories in Cityscapes and BDD100K.}
\label{tab:fid_classes}
\begin{threeparttable}
\resizebox{.8\textwidth}{!}{%
\begin{tabular}{l|rrrrrrr}
\toprule
\textbf{FID (↓)} & \textbf{car} & \textbf{bus} & \textbf{person} & \textbf{bicycle} & \textbf{motor} & \textbf{truck} & \textbf{train} \\
\midrule
Cityscapes & 145.47 & 223.02 & 211.11 & 150.94 & 207.15 & 247.85 & 162.74 \\
BDD100K   & 134.27& 150.01 & 186.33 &164.02 & 150.12 & 126.78 & 228.94 \\
\bottomrule
\end{tabular}
} % resize

\end{threeparttable}
\end{table}

In addition to the qualitative results, we further evaluate the quality of the generated images using the Fréchet Inception Distance (FID), as summarized in Table~\ref{tab:fid}. We compare three generation strategies: (1) layout diffusion~\cite{zheng2023layoutdiffusion}, which enables controllable spatial layout for object placement; (2) a two-stage pipeline that combines layout diffusion with a fast style transfer module~\cite{johnson2016FastNeuralStyleTransferNetwork}; and (3) our proposed target-aware diffusion, which directly synthesizes specific object categories guided by prompts.

The results show that our target-aware diffusion module achieves significantly lower FID values compared to the original target data (Cityscapes dataset), indicating higher realism and closer alignment to the target-domain distribution. For our method, the reported FID is the average across all object categories, whereas other methods cannot compute category-wise averages because they do not generate class-specific data. The detailed FID scores for individual categories are presented in Table~\ref{tab:fid_classes}, which includes both Cityscapes and BDD100K. We observe that categories such as car, bicycle, and train exhibit better generation quality—consistent with the qualitative results shown in Figure~\ref{fig:bddgen_example}—while person shows relatively poor quality due to facial blurring artifacts, leading to higher FID values.

Overall, since the quantitative results indicate that the target-aware diffusion approach produces more realistic and domain-aligned images than localization-controlled methods (e.g., layout diffusion and two-stage generation), we adopt the target-aware diffusion-based generation to enhance long-tailed categories as our primary data augmentation strategy.
\color{black}

\subsubsection{Comparison of Model Fusion Strategies}
\label{sub:ablation_model_fusion}

\begin{table}[htbp]
\centering
\caption{Ablation Study: $\Delta$ AP50 = FedMA - FedAvg across two datasets}
\label{table:Fedma_all}
%\makebox[\textwidth][c]{%
\begin{threeparttable}
\resizebox{\textwidth}{!}{%
\begin{tabular}{ccccc|rrrrrrrr|c}
\toprule
\multirow{2}{*}{\textbf{Dataset}} & 
\multirow{2}{*}{\textbf{Model}} & 
\multicolumn{3}{c|}{\textbf{Components}} & 
\multicolumn{8}{c|}{\textbf{$\Delta$ AP50}} & 
\multirow{2}{*}{\textbf{$\Delta$ mAP50}} \\
\cmidrule(lr){3-5}
\cmidrule(lr){6-13}
 & & $\mathcal{L}_p$ & $\mathcal{L}_{kd}$& $\mathcal{L}_{moon}$& \textbf{car} & \textbf{truck} & \textbf{rider} & \textbf{person} & \textbf{motor*} & \textbf{bicycle*} & \textbf{bus*} & \textbf{train} & \\
\midrule
\multirow{4}{*}{\textit{CK} $\rightarrow$ B} 
    & local (\textit{C})  & \checkmark& &    & -0.28  & 0.25   & -0.19  & 4.13  & 0.19  & -4.40  & 1.78   & \textemdash   & 0.21 \\
    & global & \checkmark&\checkmark&     & -0.92  & -1.12  & -1.16  & 0.91  & 3.42  & -1.65  & -1.09  & \textemdash   & -0.23 \\
    & local (\textit{C})  & \checkmark& &   \checkmark & -0.67  & 2.03   & -0.88  & 1.00  & 0.11  & 0.64   & -0.18  & \textemdash   & 0.29 \\
    
    & global & \checkmark&\checkmark&   \checkmark  & 1.51   & 1.45   & 2.09   & 4.63  & 5.02  & 2.37   & 0.31   & \textemdash   & 2.48 \\
\midrule
\multirow{4}{*}{\textit{SKF} $\rightarrow$ C} 
    & local (\textit{F})  & \checkmark& &     & -0.40  & -1.30  & -1.93  & -3.22 & -8.34 & -4.63  & -7.65  & -8.91  & -4.55 \\
    & global & \checkmark& \checkmark&     & -0.16  & -3.55  & -1.76  & -0.17 & -9.82 & -3.39  & -12.71 & -14.69 & -5.78 \\
    & local (\textit{F}) & \checkmark&&   \checkmark  & -0.17  & -6.13  & -0.82  & -4.48 & -4.88 & -5.96  & -8.56  & -2.71  & -4.21 \\
    
    & global & \checkmark& \checkmark&  \checkmark  & -0.58  & -7.95  & -1.64  & -1.12 & -9.79 & -2.01  & -22.61 & -14.17 & -7.48 \\
\bottomrule

\end{tabular}
} %resize
\begin{tablenotes}
\scriptsize
\item[*] Categories marked with a star (*) indicate object classes that are not available in the KITTI dataset.
\end{tablenotes}
\end{threeparttable}
%} %makebox
\end{table}
\noindent
\textbf{FedMA Comparison.}
In this section, we investigate the effectiveness of different backbone fusion algorithms: FedAvg \cite{FedAvg} and FedMA \cite{FedMA}. Table~\ref{table:Fedma_all} presents the performance differences between these two methods across different experimental settings. For the local models, we report AP results on Cityscapes in the \textit{CK→B} setting and Foggy Cityscapes in the \textit{SKF→C} setting, as both datasets cover all categories. In terms of overall mAP50, FedMA generally leads to improvements in the \textit{CK→B} scenario. However, in the \textit{SKF→C} setting, it consistently results in performance drops across nearly all categories. We hypothesize that although Foggy Cityscapes and Cityscapes share similar object layouts, the presence of fog significantly alters the style distribution. This discrepancy may lead to large variations in the features extracted by the backbone, and during the layer-wise matching of FedMA, features from the foggy domain may dominate the fusion process, ultimately degrading the global model.

Moreover, even in the \textit{CK→B} setting, the benefits of FedMA are not consistent across all categories. For the local results, we report the performance of the Cityscapes model since it contains annotations for all eight categories, whereas KITTI lacks several such as \textit{motorcycle}, \textit{bicycle}, and \textit{bus}, which are denoted with an star. However, performance varies between categories, with some showing improvement and others declining. These changes do not appear to correlate with whether a category is shared among all clients or exclusive to one, indicating that FedMA is less stable than FedAvg.

\begin{table*}[htbp]
\centering
\caption{\color{blue}Comparison of different model fusion strategies, FedAvg and FedProx, under \textit{CK$\rightarrow$B} and \textit{SKF$\rightarrow$C}.\color{black}}
\label{tab:fedprox_comparison}
\resizebox{\textwidth}{!}{%

\begin{tabular}{cccc|ccccccc|c}
\toprule[1.5pt]

\multirow{2}{*}{\textbf{Dataset}} & 
\multirow{2}{*}{\textbf{Name}} & 

\multicolumn{2}{c|}{\textbf{Components}} & 
\multicolumn{7}{c|}{\textbf{AP}} & 
\multirow{2}{*}{\textbf{ mAP}} \\
\cmidrule(lr){3-4}
\cmidrule(lr){5-11}

 & & Model Fusion & Gen.&  \textbf{car} & \textbf{truck} & \textbf{rider} & \textbf{person} & \textbf{motor} & \textbf{bicycle} & \textbf{bus}  & \\

\midrule
\multirow{6}{*}{CK$\rightarrow$B} 
& FedCoin & FedAvg & \textemdash   & 52.40 & 22.20 & 35.50 & 35.50 & 19.10 & 29.10 & 25.10 & 31.30\phantom{($\downarrow$)} \\
&  \textemdash       & FedProx & \textemdash  & 50.30 & 21.59 & 30.89 & 36.85 & 15.38 & 28.30 & 21.90 & 29.32 ($\downarrow$) \\
\cmidrule(lr){2-12}

& HeroCrystal-all & FedAvg & all  & 49.70 & 27.50 & 35.40 & 36.20 & 21.10 & 29.90 & 27.50 & 32.47 \phantom{($\downarrow$)}\\
&     \textemdash              & FedProx & all  & 42.98 & 26.78 & 33.14 & 37.37 & 19.35 & 26.12 & 25.65 & 30.20  ($\downarrow$) \\
\cmidrule(lr){2-12}

& HeroCrystal-rare & FedAvg & rare & 52.80 & 27.70 & 36.10 & 37.20 & 21.60 & 31.20 & 27.70 & 33.47\phantom{($\downarrow$)} \\
&      \textemdash             & FedProx & rare & 53.48 & 26.73 & 33.89 & 37.07 & 20.79 & 30.23 & 26.85 & 32.72 ($\downarrow$)  \\
\midrule[1.5pt]
\multirow{6}{*}{SKF$\rightarrow$C} 
& FedCoin & FedAvg & \textemdash   & 62.70 & 32.00 & 45.30 & 45.30 & 30.40 & 39.50 & 50.80 & 42.50 \phantom{($\downarrow$)}\\
&  \textemdash      & FedProx & \textemdash   & 62.73 & 28.69 & 35.87 & 39.19 & 28.99 & 39.52 & 55.90 & 41.56  ($\downarrow$) \\
\cmidrule(lr){2-12}

& HeroCrystal-all & FedAvg & all  & 65.20 & 40.20 & 42.70 & 40.20 & 32.00 & 40.80 & 58.30 & 45.62 \phantom{($\downarrow$)}\\
&      \textemdash             & FedProx & all  & 63.54 & 31.44 & 44.38 & 39.24 & 30.98 & 35.39 & 49.16 & 42.02 ($\downarrow$)  \\
\cmidrule(lr){2-12}

& HeroCrystal-rare & FedAvg & rare & 65.70 & 36.10 & 38.40 & 39.70 & 32.20 & 43.30 & 63.60 & 45.57 \phantom{($\downarrow$)}\\
&      \textemdash             & FedProx & rare & 62.28 & 25.99 & 37.81 & 40.01 & 32.02 & 38.29 & 52.27 & 41.24  ($\downarrow$) \\
\bottomrule[1.5pt]

\end{tabular}
}
\end{table*}

\color{blue}

\noindent
\textbf{FedProx Comparison.} In addition, we provide a comparison with FedProx \cite{li2020fedprox}, a widely used extension of FedAvg that introduces a proximal term to restrict local updates from deviating too far from the global model.\footnote{Following the original FedProx configuration, the proximal coefficient is set to $\mu = 0.01$.} As shown in Table~\ref{tab:fedprox_comparison}, we further replace the default FedAvg strategy with FedProx across all three methods, FedCoin, HeroCrystal-all, and HeroCrystal-rare, under both \textit{CK$\rightarrow$B} and \textit{SKF$\rightarrow$C} settings. The results indicate that adopting FedProx does not provide any performance gain; in fact, it often leads to a slight drop in accuracy. We hypothesize that this is because our framework already employs dynamic model contrastive learning (Sec.~\ref{subsub:dynamic_moon}), which was specifically designed to address the observation that global models tend to underperform in early rounds. By encouraging local models to retain more influence during the initial training phase, this strategy alleviates the sharp accuracy decline we observed at the beginning of training. In contrast, FedProx enforces the opposite principle: it regularizes local updates to remain close to the global model, thereby constraining the degree to which local representations can deviate. This regularization undermines the benefits of our contrastive mechanism, which relies on allowing local models to preserve more domain-specific features in the early stage. Consequently, integrating FedProx suppresses the intended effect of our dynamic contrastive learning, leading to reduced performance.

Our experiments show that when dynamic model contrastive learning is employed to stabilize local–global representation alignment, replacing FedAvg with FedProx provides no clear benefit and even suppresses the desired effect of leveraging richer local representations. On the other hand, FedMA introduces instability and substantially higher communication overhead without consistent performance improvement (see Sec.~\ref{subsub:fedma_communication} for detailed communication cost analysis). Therefore, we adopt the simpler and more stable FedAvg strategy in our final framework, which achieves effective model fusion without incurring additional computational cost.

\color{black}

\subsubsection{Impact of Probabilistic Modeling on Pseudo-label Accuracy}
\label{sub:FN_reduce_ablation}

\begin{figure}
\centering

\subfloat[]{
\includegraphics[width=.7\linewidth]{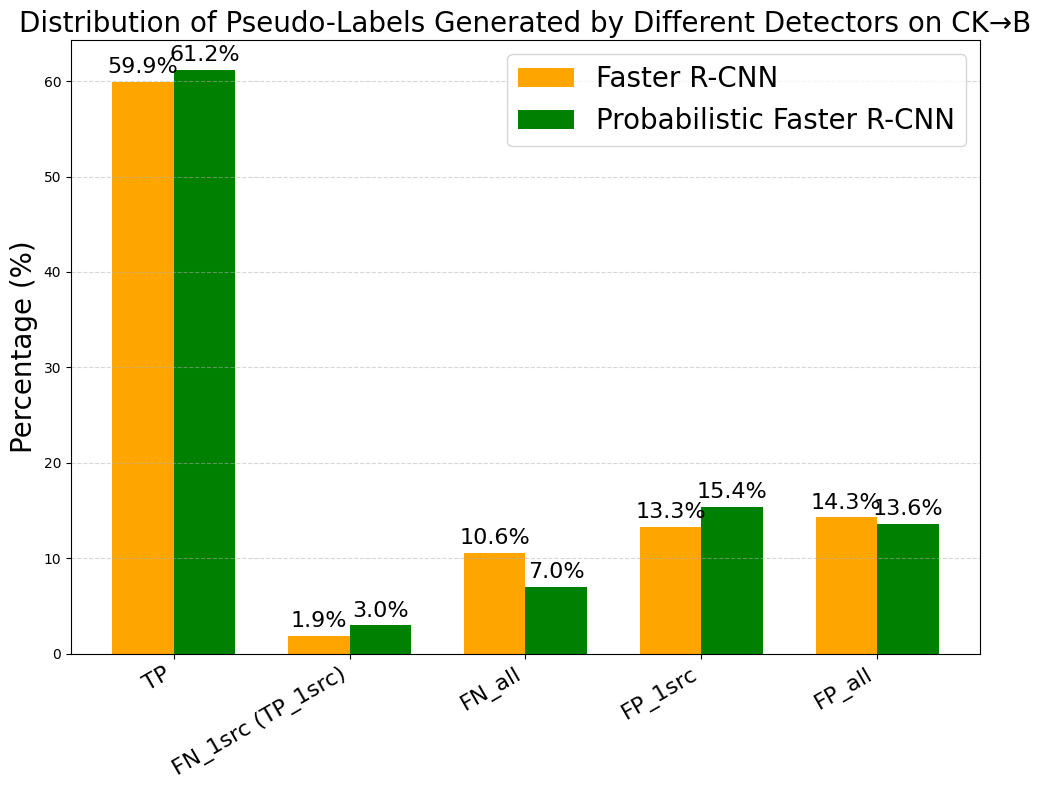}}
\\
\subfloat[]{
\includegraphics[width=.9\linewidth]{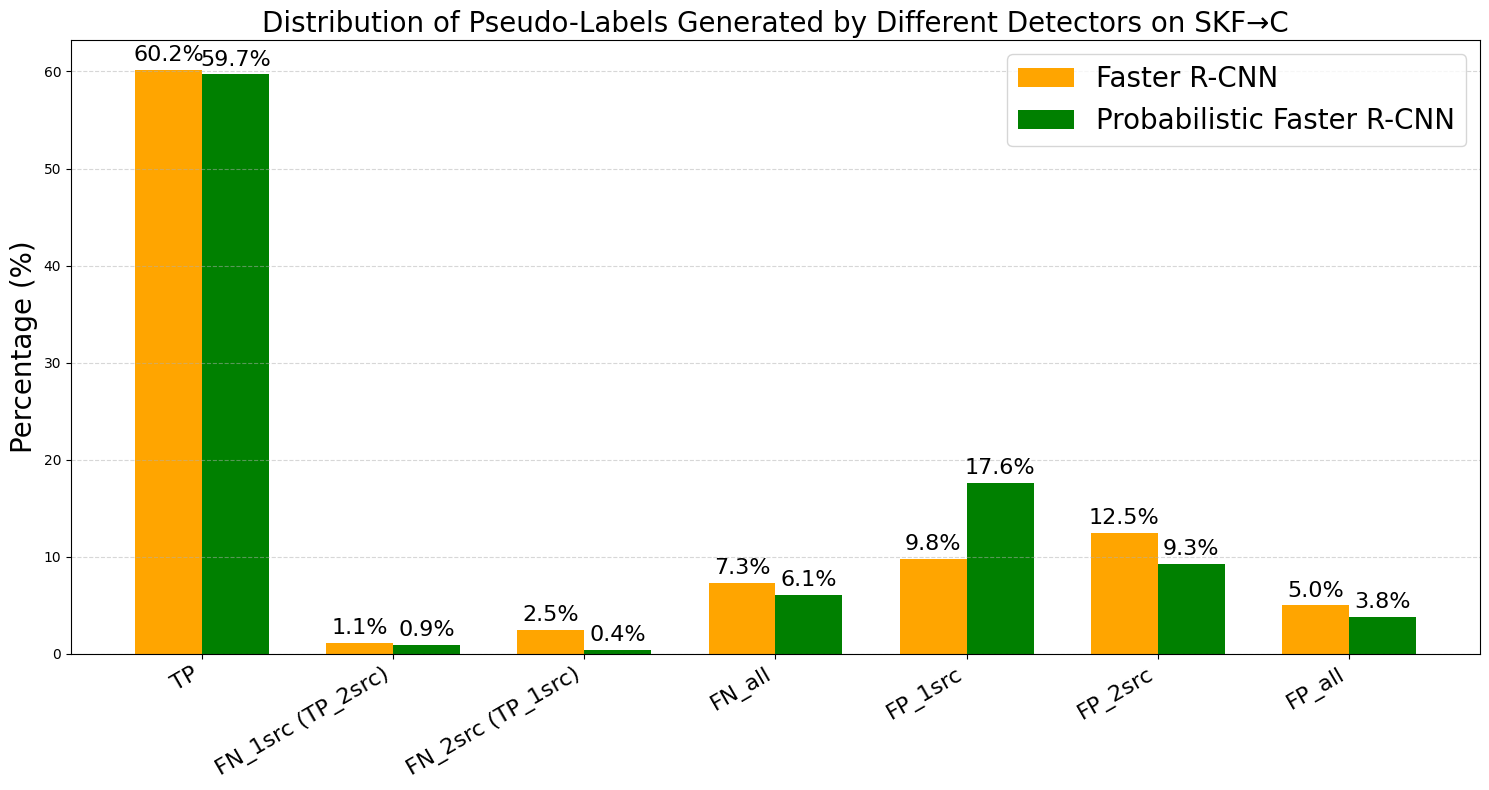}}
\caption{Distribution of pseudo-labels generated by Faster R-CNN and Probabilistic Faster R-CNN on (a) \textit{CK$\rightarrow$B} and (b) \textit{SKF$\rightarrow$C}. Bars indicate the proportion of true positives (TP), false negatives (FN), and false positives (FP), highlighting differences in pseudo-label quality.}
\label{fig:pseudo-label_trend}
\end{figure}

Figure~\ref{fig:pseudo-label_trend} (a) and (b) show the distribution of pseudo-labels generated by Faster R-CNN and Probabilistic Faster R-CNN (Sec. \ref{subsub:PFRCNN}) on \textit{CK$\rightarrow$B} and \textit{SKF$\rightarrow$C}, respectively. These results are used to examine the changes in the quality composition of pseudo-labels—specifically the proportions of true positives (TP), false negatives (FN), and false positives (FP)—after applying Probabilistic Faster R-CNN. In the multi-source setting, \textit{FN\_$1src$} (or \textit{TP\_$1src$}) indicates that one of the source detectors failed (or succeeded) in detection, while others provided the correct prediction. In the three-source case of \textit{SKF$\rightarrow$C}, \textit{FN\_$2src$} and \textit{TP\_$2src$} represent that two sources failed or succeeded, respectively.

Notably, Probabilistic Faster R-CNN significantly reduces \textit{FN\_all} (e.g., from 10.6\% to 7.0\% in \textit{CK$\rightarrow$B}) and improves TP, but also introduces more \textit{FP\_$1src$} (e.g., +2.1\% and +7.8\% in \textit{CK$\rightarrow$B} and \textit{SKF$\rightarrow$C}, respectively). These increased false positives are later addressed by our proposed ICI algorithm, which integrates cross-source consensus to refine the final pseudo-labels.

\color{blue}
\subsection{Privacy and Ethical Analysis}
\label{sub:information_leak}
\color{black}

\begin{table}[ht]
\centering
\caption{\color{blue}Information leakage assessment using three evaluation methods (SSIM, LPIPS, and PSNR) to quantify similarity between generated images and source training images.\color{black}}
\label{tab:leakage-risk}

\resizebox{\textwidth}{!}{%
\begin{tabular}{lccc cccccccc}
\toprule
\multirow{2}{*}{\textbf{Metric}} 
  & \multicolumn{3}{c}{\textbf{Risk level}} 
  & \multicolumn{6}{c}{\textbf{Generated Category}} \\
\cmidrule(lr){2-4}\cmidrule(lr){5-11}
& \textbf{low risk} & \textbf{middle risk} & \textbf{high risk}
& \textbf{car} & \textbf{bus} & \textbf{motor} & \textbf{person} & \textbf{truck} & \textbf{bicycle} & \textbf{Avg.} \\
\midrule
SSIM  & \textbf{$>0.2\text{--}0.4$}   & $0.1\text{--}0.2$    & $\le 0.05\text{--}0.1$
      & 0.46 & 0.43 & 0.40 & 0.38 & 0.44 & 0.36 & \textbf{low risk} \\
LPIPS & \textbf{$<0.5\text{--}0.8$}   & $0.8\text{--}0.9$    & $\ge 0.9$
      & 0.57 & 0.63 & 0.61 & 0.59 & 0.67 & 0.58 & \textbf{low risk}\\
PSNR  & \textbf{$<25\text{--}30$}     & $30\text{--}35$      & $\ge 35\text{--}40$
      & 11.76 & 11.86 & 12.01 & 11.71 & 10.15 & 10.97 & \textbf{low risk}\\
\bottomrule
\end{tabular}
}
\end{table}

\color{blue}
To empirically evaluate the privacy guarantees of our one-shot, target-aware synthetic data generation, we employ three complementary evaluation metrics, such as SSIM, LPIPS, and PSNR, to quantify potential information leakage. In Table \ref{tab:leakage-risk}, we report the risk levels of each metric in different object categories. All similarity scores fall within the low-risk range (SSIM between 0.2 and 0.4, LPIPS between 0.5 and 0.8, and PSNR below 25 dB), indicating negligible likelihood of pixel-level or structural leakage from target domains. This confirms that the learned personalization token primarily captures high-level style cues rather than memorizing individual content.
Although these values indicate low risk, they also suggest a moderate degree of similarity—sufficient to retain style consistency without reproducing identifiable details. From an ethical perspective, the proposed one-shot generation avoids direct use of sensitive surveillance imagery by learning global-style statistics from a single anonymized reference, thereby minimizing privacy exposure and reducing dataset collection risks. As also illustrated in Figure \ref{fig:bddgen_example}, the human figures generated preserve recognizable category-level semantics (e.g., “\textit{person}”) while eliminating personally identifiable features such as facial details.
In summary, these privacy and ethical analyzes demonstrate that our method effectively transfers domain style and generates category-controllable data without leaking private content, ensuring both privacy preservation and ethical compliance.

\color{black}

\color{blue}
\subsection{Communication and Training Overhead Analysis}
\label{subsec:communication}

To quantify the efficiency of our framework, we analyze both computational and communication overheads. Sec.~\ref{subsub:diffusion_overhead} analyzes the computational cost introduced by the one-shot diffusion-based generation, while Sec.~\ref{subsub:fedma_communication} examines the communication overhead in federated fusion by comparing different model fusion strategies.

\subsubsection{Computational Overhead of Diffusion-based Generation}
\label{subsub:diffusion_overhead}

In the \textbf{generated stage}, additional computational resources are required for both training and inference of the diffusion-based generation module. During one-shot personalization, fine-tuning the diffusion model for a single target domain takes approximately 28–30 minutes for 1,000 training steps on a single NVIDIA V100 GPU. Once personalized, image synthesis becomes highly efficient—each image can be generated in about 2.6 seconds. Since we generate 100 target-style images for each category \textit{C}, the total overhead is $2.6\times100 \times|C|$. During the \textbf{federated stage}, an additional set of \( |C| \times 100 \) synthetic target-style images was introduced to supplement the local training data for each client.

Although the inclusion of diffusion-generated images increases the total number of training samples, the impact on the overall training time is marginal. In practice, adding 100 synthetic images per category corresponds to less than a 10\% increase in data volume, which typically leads to only a 5–8\% increase in training time. This effect is sublinear because GPU utilization and I/O overhead remain largely unchanged during batch processing. Moreover, since image generation is performed offline before training, it does not introduce additional time within each training round.

\color{blue}
\subsubsection{Communication Cost of Federated Fusion}
\label{subsub:fedma_communication}
\color{black}

\begin{figure}[htb]
\centering

   \includegraphics[width=.6\linewidth]{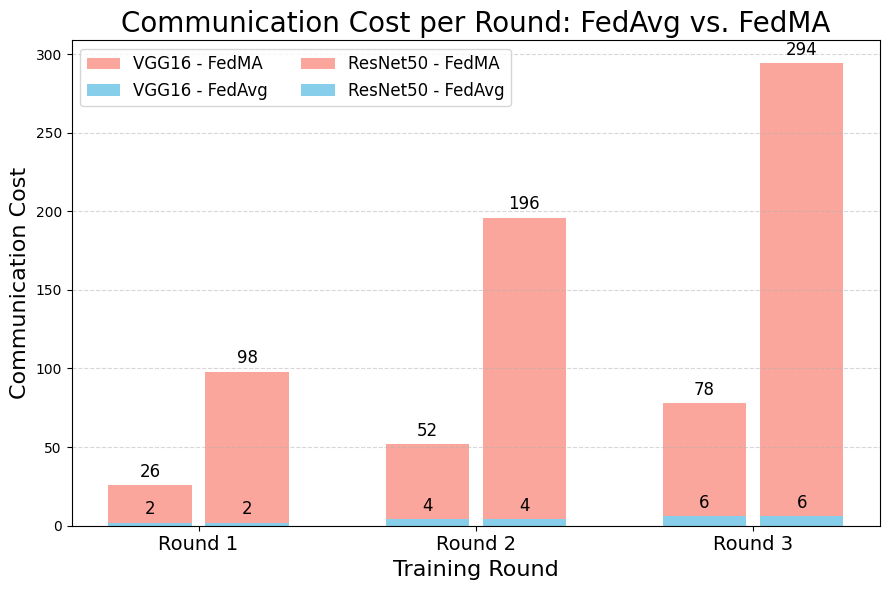}

  \caption{Communication costs of FedAvg and FedMA over training rounds using VGG16 and ResNet50 backbones.
}
  \label{fig:comm_cost}
\end{figure}

\color{blue}

In addition to the performance variability mentioned in Sec.~\ref{sub:ablation_model_fusion}, \color{black} we also compare the communication costs between FedAvg and FedMA under different backbone architectures. As shown in Figure~\ref{fig:comm_cost}, FedAvg transmits the global model twice per round, leading to a predictable and fixed cost. In contrast, FedMA involves additional communication overhead due to its layer-wise matching process, which increases with the number of unfrozen layers. This effect becomes more pronounced when using deeper backbones such as ResNet50. Therefore, while FedMA may offer alignment advantages in theory, it introduces substantial communication inefficiency in practice.

\subsection{Qualitative Results}
\label{subsec:qualitative}

\begin{figure}[htbp]
  \centering
  \begin{minipage}[t]{0.49\linewidth}
    \centering
    \includegraphics[width=\linewidth]{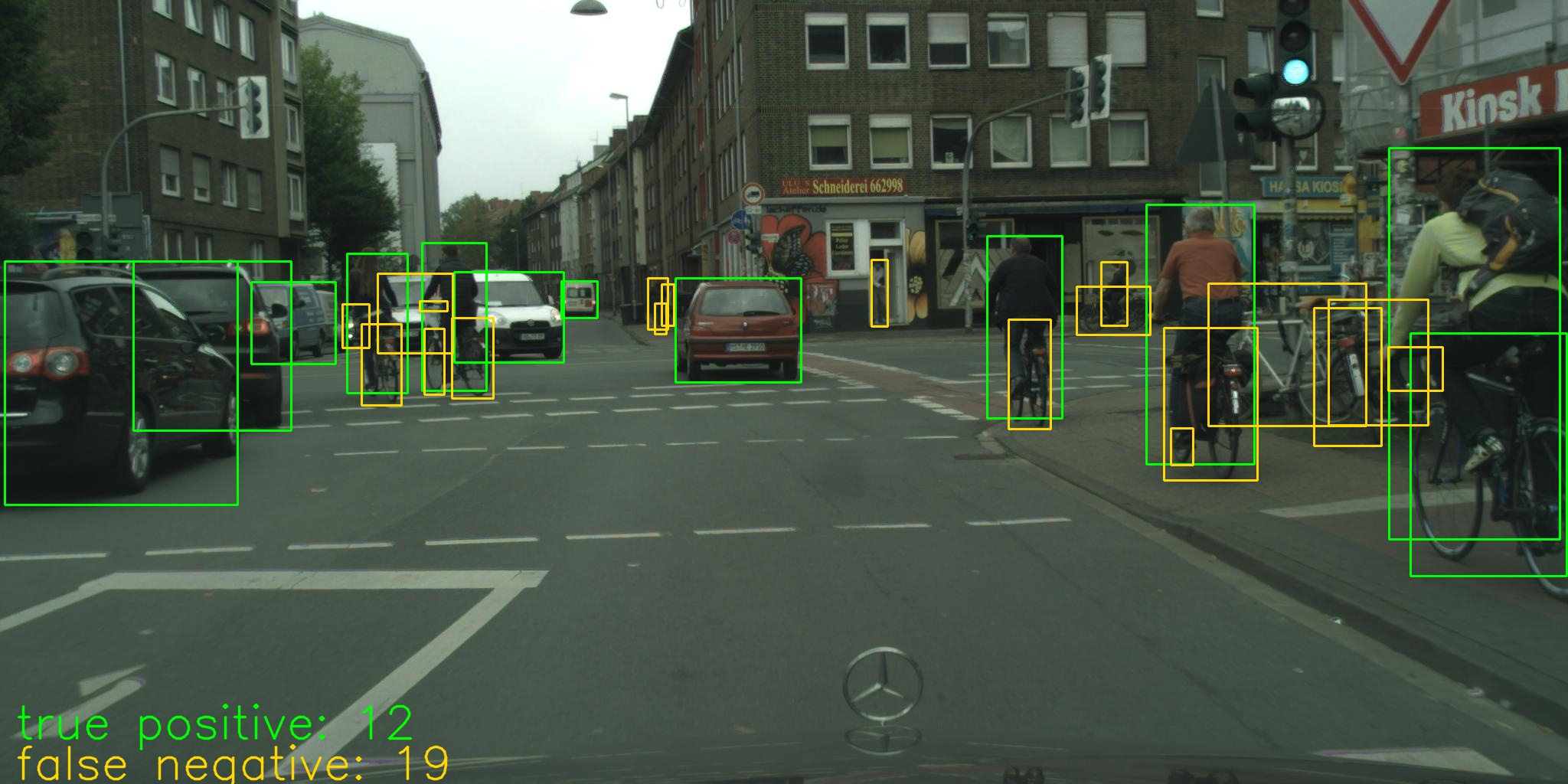}
    \subfloat[]{}
  \end{minipage}
  \hfill
  \begin{minipage}[t]{0.49\linewidth}
    \centering
    \includegraphics[width=\linewidth]{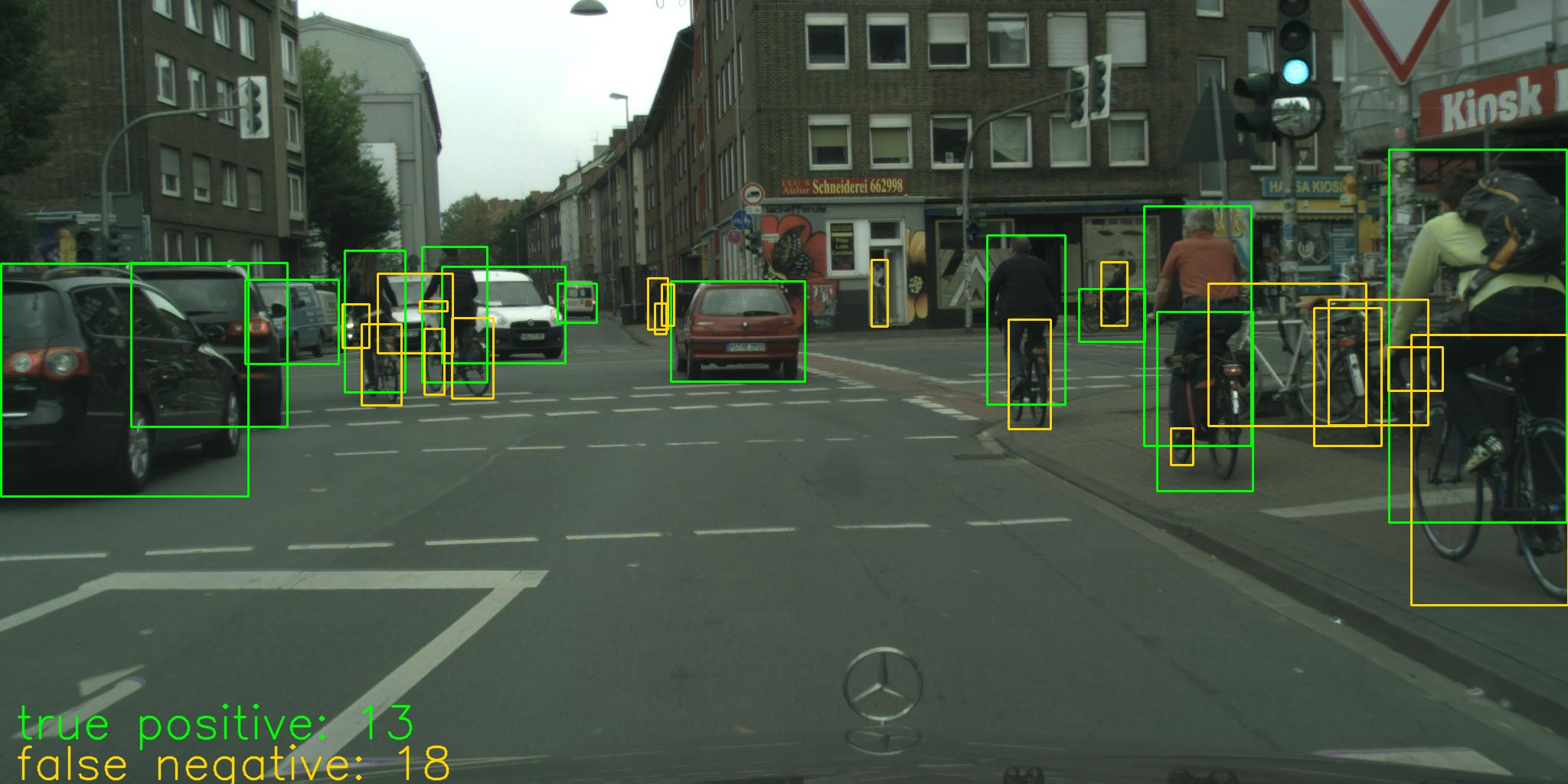}
    \subfloat[]{}
  \end{minipage}
  \\[4pt]
  \begin{minipage}[t]{0.49\linewidth}
    \centering
    \includegraphics[width=\linewidth]{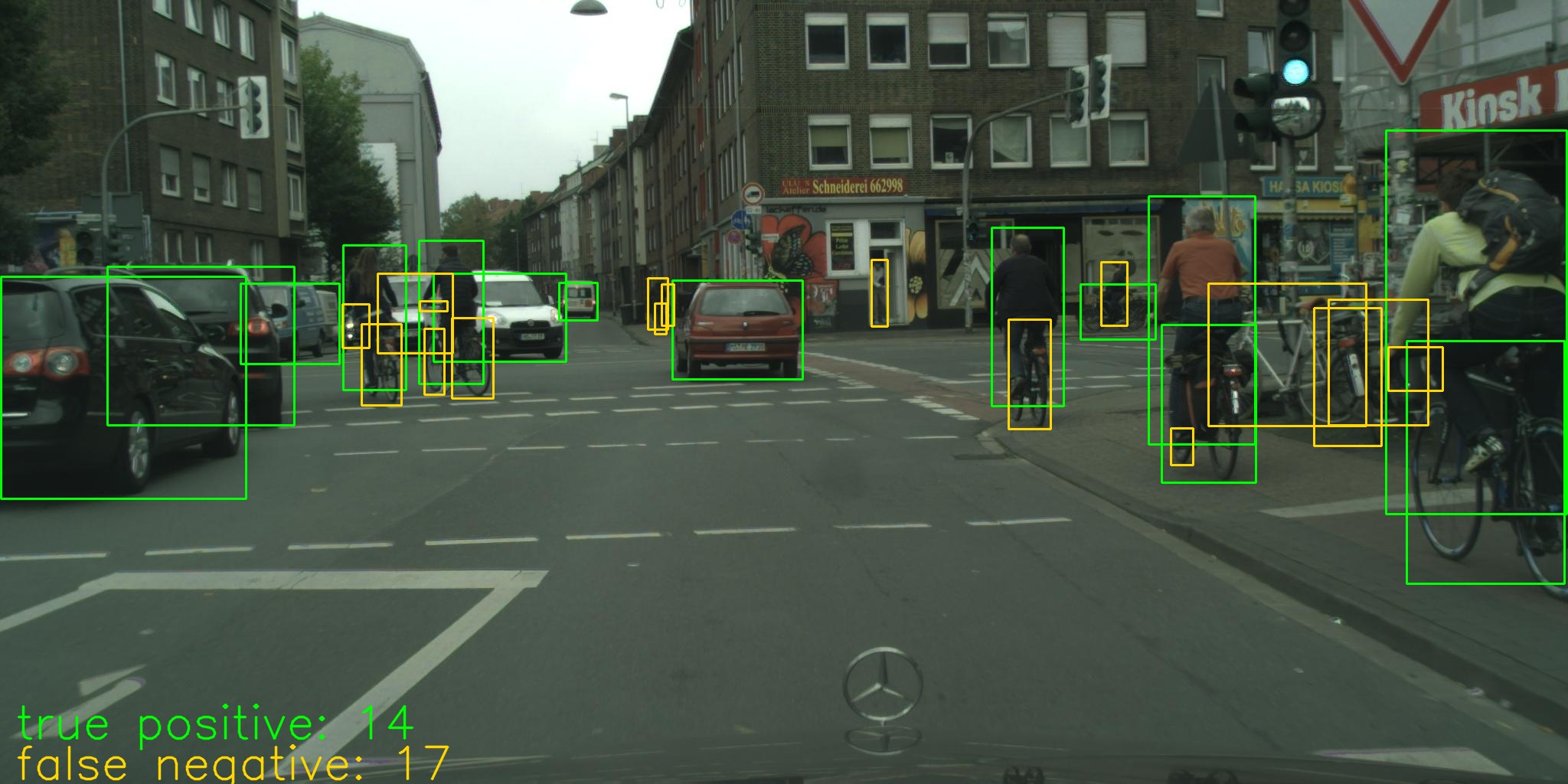}
    \subfloat[]{}
  \end{minipage}
  \hfill
  \begin{minipage}[t]{0.49\linewidth}
    \centering
    \includegraphics[width=\linewidth]{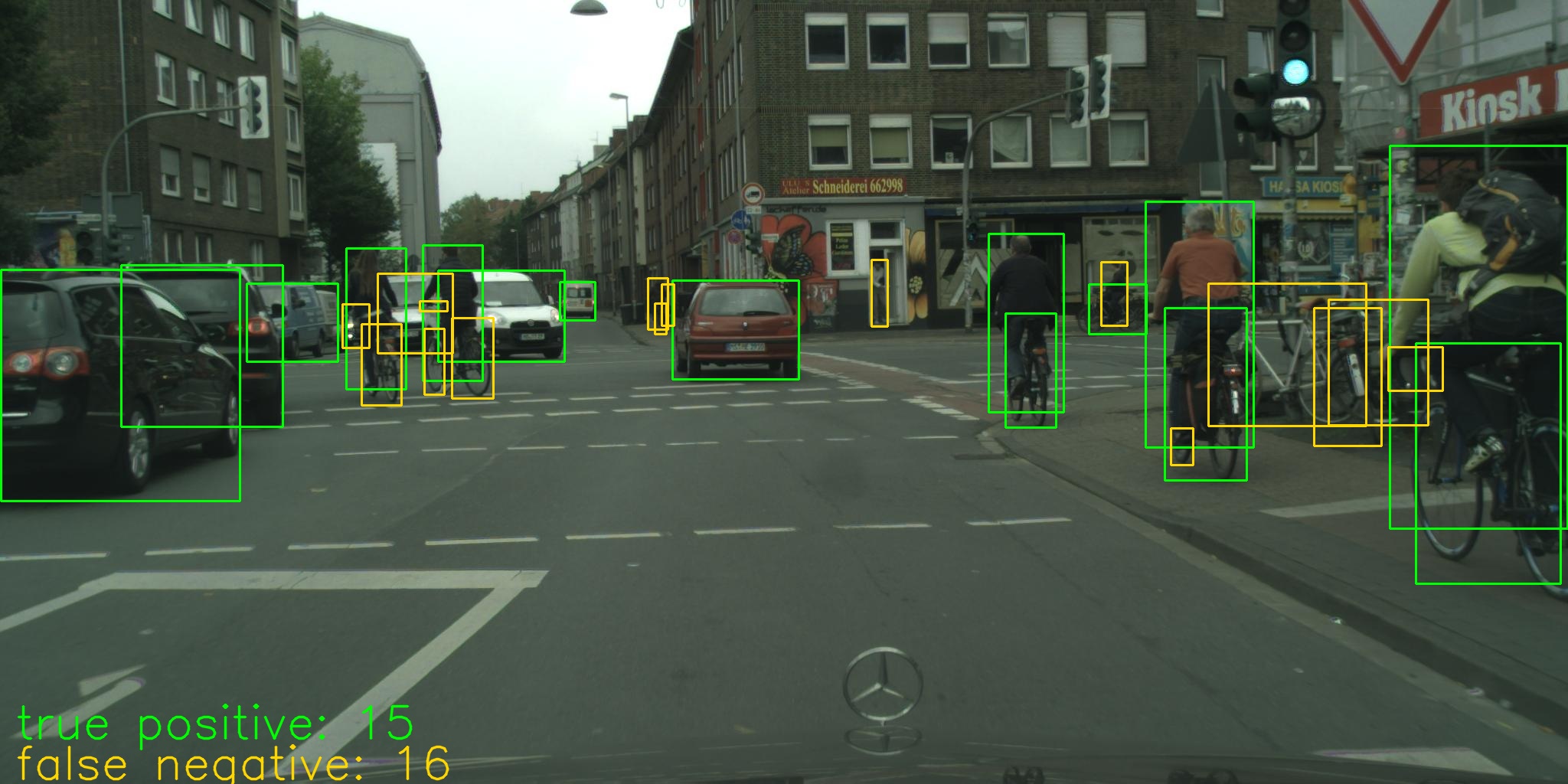}
    \subfloat[]{}
  \end{minipage}
  \caption{The effect of using ICI on \textit{SKF$\rightarrow$C} for (a) FedAvg, (b) FedMA, (c) FedCoin, and (d) HeroCrystal, where green boxes represent true positives and yellow boxes represent false negatives.} 
  \label{fig:qualitative_results}
\end{figure}

This study presents a qualitative analysis of the performance variations of several algorithms following the application of the ICI algorithm. Figure \ref{fig:qualitative_results} illustrates the object detection outcomes for the \textit{SKF$\rightarrow$C} example using these algorithms: FedAvg achieved 12 true positives and 19 false negatives; FedMA recorded 13 true positives and 18 false negatives; FedCoin obtained 14 true positives and 17 false negatives; and HeroCrystal attained the highest performance with 15 true positives and 16 false negatives. The findings suggest that FedAvg exhibits multi-target detection capability under the influence of the ICI algorithm, while HeroCrystal demonstrates the best overall performance with the greatest number of true positives and the fewest false negatives.

\section{Conclusion and Future Works}
This work identifies key challenges in multi-camera domain-adaptive object detection, including long-tailed category imbalance, privacy constraints, and heterogeneous model architectures. To address these issues, we propose \textit{HeroCrystal}, a unified framework that combines target-aware image generation with FL. 
% waue revised 20251003
\color{blue}
Our generation module adapts to target-domain styles and enables controllable synthesis of specific object categories, helping to reduce rare-class performance drops. The proposed ICI algorithm achieves privacy-preserving fusion of heterogeneous models in a federated setting. Extensive experiments and ablation studies validate the effectiveness of each component.
\begin{comment}

Our generation module not only learns target-domain styles from a single image but also enables controllable synthesis of specific object categories, effectively mitigating the performance drop in rare classes. Furthermore, the proposed ICI algorithm facilitates privacy-preserving fusion of heterogeneous models within a federated architecture. Extensive experiments across diverse settings validate the effectiveness of our approach, and ablation studies provide detailed insights into the role of each component.
\end{comment}

%\begin{itemize}
    %\item \textbf{Computation Overhead}: Additional resource demands arise from training the diffusion model, generating images via the diffusion module, and synthesizing new images per training iteration.
 %   \item \textbf{Future Work}: 
 In the future, we plan to further enhance semantic diversity among dominant object categories during generation, aiming to boost overall detection accuracy while preserving the gains in rare-class augmentation. Moreover, we will explore integrating spatially controllable generation frameworks with our target-aware diffusion module to combine spatial control, style transfer, and category-specific generation, further improving long-tailed category performance and domain adaptation accuracy. \color{red}Additionally, we plan to explore extending our framework to one-stage and transformer-based detectors, which would require redesigning the backbone–head decomposition and probabilistic formulation for alternative detection paradigms. \color{black}

%\end{itemize}
\color{black}

\section*{Acknowledgment}
This work was supported by the National Science and Technology Council (NSTC), Taiwan, under grant number NSTC 113-2222-E-194-003 and 114-2221-E-194-030-MY3. We thank the National Center for High-performance Computing (NCHC) for providing computational and storage resources.
\clearpage

\begingroup
\small
\bibliography{refs}
\endgroup
\end{document}